\PassOptionsToPackage{table}{xcolor}
\documentclass[runningheads]{llncs}

\usepackage{eccv}

\usepackage{eccvabbrv}

\usepackage{graphicx}
\usepackage{booktabs}
\usepackage{multirow}
\usepackage{pifont}
\usepackage{xcolor}
\usepackage{capt-of}
\usepackage{placeins}
\usepackage{docmute}
\definecolor{gray}{gray}{0.4}

\usepackage[accsupp]{axessibility}  

\usepackage{hyperref}

\usepackage{orcidlink}

\begin{document}

\def\lastandname{,}

\title{Reflect-R1: Evidence-Driven Reflection for Self-Correction in Long Video Understanding} 

\titlerunning{Reflect-R1}

\author{Shuimu Chen\inst{1}$^*$ \and
Yuteng Chen\inst{3}$^*$ \and
Yuanshen Guan\inst{4}$^*$ \and
Zebang Cheng\inst{2,5}$^\dagger$ \and
Zeyu Zhang\inst{6} \and
Shengqian Qin\inst{7} \and
Bin Xia\inst{1} \and
Jiaran Li\inst{1} \and
Wenming Yang\inst{1}$^\ddagger$ \and
Fei Ma\inst{2}$^\ddagger$}

\authorrunning{S.~Chen et al.}

\institute{%
{\fontsize{6.8pt}{7.5pt}\selectfont
\begin{tabular}{@{}c@{}}
\inst{1} Tsinghua University \hspace{0.8em}
\inst{2} Guangdong Laboratory of Artificial Intelligence and Digital Economy (SZ)\\
\inst{3} Nanyang Technological University \hspace{0.8em}
\inst{4} University of Science and Technology of China\\
\inst{5} Shenzhen University \hspace{0.8em}
\inst{6} University of California \hspace{0.8em}
\inst{7} Shanghai Jiao Tong University
\end{tabular}}}

\maketitle
\noindent\makebox[\textwidth][c]{%
{\fontsize{7.2pt}{8pt}\selectfont
\textbf{Project:} \href{https://github.com/ShuimuChen-hyq/Reflect-R1}{https://github.com/ShuimuChen-hyq/Reflect-R1}}}
\par
\begingroup
\renewcommand{\thefootnote}{*}
\footnotetext{Equal contribution.}
\renewcommand{\thefootnote}{$\dagger$}
\footnotetext{Project Lead.}
\renewcommand{\thefootnote}{$\ddagger$}
\footnotetext{Corresponding author.}
\endgroup


\begin{abstract}
Current multimodal reflection mechanisms for long video understanding predominantly rely on closed-loop self-reflection within internal parameters. Lacking objective external evidence, models are frequently trapped in blind confidence and often fail to correct errors. Furthermore, applying reinforcement learning to multi-stage reflection pipelines introduces severe policy coupling, which is exacerbated by a critical scarcity of dedicated training data. To address these limitations, this work proposes Reflect-R1, the first Evidence-Driven self-correction framework for long video understanding. The framework constructs a three-stage pipeline consisting of intuition, verification, and arbitration. By dynamically retrieving objective visual evidence to verify initial intuitions and autonomously executing multiple temporal searches to resolve conflicts, it completely breaks the hallucination loop. To overcome policy coupling, we design a stage-decoupled reinforcement learning algorithm named SD-GRPO that independently computes advantage functions across different reasoning stages. Concurrently, we construct a dataset of 120K samples to bridge the training data gap. Extensive experiments on benchmarks such as VideoMME and LongVideoBench demonstrate that Reflect-R1 achieves state-of-the-art performance. Our method significantly improves the genuine rectification rate and enables authentic self-correction strictly grounded in objective evidence.
\end{abstract}




\begin{figure}[t]
\centering
\includegraphics[width=\textwidth]{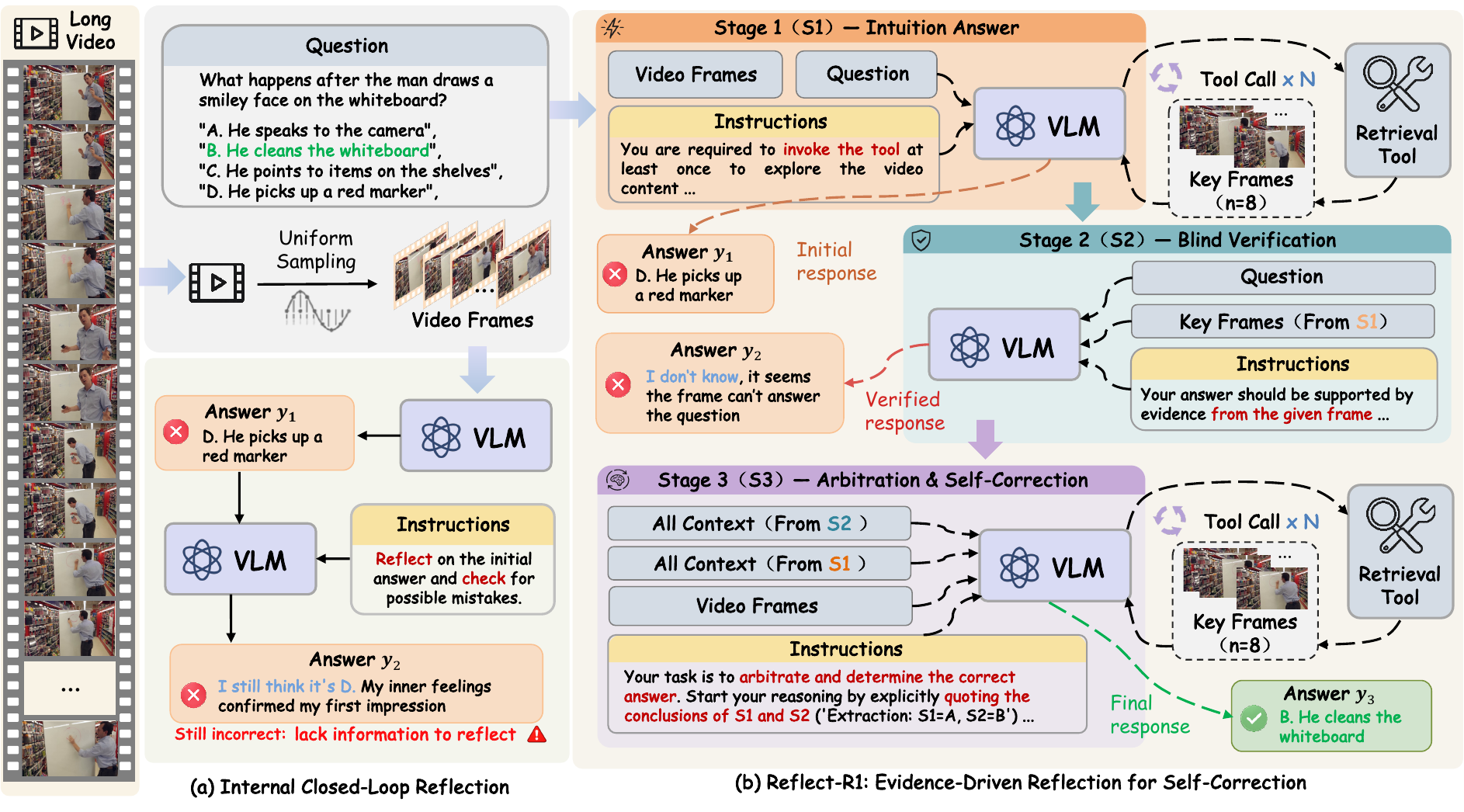}
\caption{\textbf{Comparison between Internal Closed-Loop Reflection and Evidence-Driven Reflection.} (a) Traditional closed-loop reflection relies solely on internal parametric knowledge, easily falling into the trap of blind confidence and failing to correct errors. (b) Reflect-R1 completely breaks the hallucination loop by formalizing an ``intuition-verification-arbitration'' pipeline, executing active searches to achieve genuine self-correction strictly grounded in objective retrieved evidence.}
\label{fig:fig1}
\end{figure}

\section{Introduction}
\label{sec:intro}

Long video understanding~\cite{pereira2025self,wu2024longvideobench,he2024ma,tang2025adaptive,chenzhaoyu2026inverttvg,Guo_2026_CVPR} is a critical task for applying artificial intelligence to complex real-world scenarios. Recent video-oriented multimodal large language models (MLLMs) further broaden this landscape to event streams, personalized video chat, and fine-grained facial video understanding~\cite{Liu_2025_CVPR,Shi_2025_ICCV,zhao2026favchathierarchicalpromptqueryguided}. Beyond understanding visual content, reliable deployment also requires reflection, where a model explicitly scrutinizes and potentially corrects its initial intuitive output before generating a final response. Recent pioneering works, including Vision-R1~\cite{huang2025vision}, VL-Rethinker~\cite{VL-Rethinker}, and Video-R1~\cite{feng2025video}, have advanced the field by integrating such reflection mechanisms into MLLMs to mitigate hallucinations and capture visual details. Concurrently, large language models such as OpenAI o1~\cite{openai2024openaio1card} and DeepSeek-R1\cite{Guo_2025} demonstrate that eliciting long chain-of-thought (CoT) reasoning~\cite{wei2022chain,kojima2022large,10.1145/3746027.3755837} through reinforcement learning substantially enhances reflection and complex logical deduction. However, bringing such reflection paradigms to multimodal long-video understanding is difficult. Most existing methods perform reflection in a closed-loop internal manner, which gives rise to two main problems.

The first problem is verification failure caused by a lack of objective evidence. As illustrated in Fig.~\ref{fig:fig1} (a), this paradigm forces the model to rely solely on internal knowledge to repeatedly scrutinize the initial answer ($y_1$). Because long video information is complex and independent external visual evidence is absent, this completely closed internal reasoning process easily traps the model in blind confidence~\cite{pan2026ground,kulkarni2025avatar}. When attempting to correct errors, the model frequently uses internally generated hallucinations to forcibly justify the initial erroneous conjecture, making the reflection process ineffective or even counterproductive. The empirical analysis in ~\cref{sec:Preliminary} clearly confirms this phenomenon. Without external verification, the reflection process of multimodal models often degenerates into random alterations~\cite{pan2026ground}, where the probability of changing a correct answer to an incorrect one frequently exceeds the probability of correcting errors. 

The second problem involves policy coupling~\cite{wang2025practitioner} during reinforcement learning optimization and an acute scarcity of training data. To elicit reflection capabilities through reinforcement learning, it is intuitively necessary to jointly train the initial answering phase and the subsequent correction phase within a unified trajectory. However, applying standard reinforcement learning directly to such a long-chain, multi-stage process triggers severe policy coupling. Specifically, the complexity of prolonged reasoning drives the model to exploit optimization shortcuts, such as simply repeating the initial guess during the reflection stage to secure base rewards instead of learning authentic error-correction logic~\cite{cheng2025stop}. Furthermore, the extreme lack of high-quality training data tailored for multimodal reflection remains a critical bottleneck that prevents models from developing deep self-correction abilities.

To address verification failures caused by closed-loop blind confidence, we propose Reflect-R1. To the best of our knowledge, Reflect‑R1 is the first evidence‑driven self‑correction framework for long video understanding that explicitly decomposes reflection into an intuition–verification–arbitration pipeline. In the first stage, the model generates an intuitive answer ($y_1$) and actively retrieves keyframes as external visual evidence. In the second stage, the model performs an independent blind verification ($y_2$) by relying exclusively on these retrieved frames to assess the initial intuition. This process ensures that the model evaluates the question based on objective evidence while maintaining strict information isolation from the global video context. Finally, the arbitration stage ($y_3$) resolves conflicts between the subjective intuition and the objective verification result to produce a final response. If the initial evidence is insufficient for a definitive conclusion, the model autonomously re-invokes temporal search tools until conclusive proof is captured. By grounding the entire reflection process in external visual evidence, Reflect-R1 effectively breaks the hallucination loop and achieves authentic multimodal self-correction.

Furthermore, to overcome policy coupling in GRPO~\cite{shao2024deepseekmath} and bridge the training data gap, we design a novel Stage-Decoupled GRPO (SD-GRPO) algorithm along with dedicated datasets. The SD-GRPO algorithm effectively prevents the model from seeking optimization shortcuts by computing the advantage function independently across different reasoning stages, including intuition, verification, and arbitration. This mechanism forces the model to learn genuine error correction logic. Concurrently, we systematically construct Reflect-R1-CoT-90k for supervised fine-tuning cold start and Reflect-R1-RL-30k for reinforcement learning training, fully resolving the data scarcity bottleneck in this field.

Reflect-R1 achieves state-of-the-art performance across major benchmarks including VideoMME~\cite{fu2025video}, LongVideoBench~\cite{wu2024longvideobench}, and MLVU~\cite{zhou2025mlvu}. It also demonstrates superior localization precision in temporal search tasks on Haystack-LVBench~\cite{ye2025rethinking}. Most importantly, Reflect-R1 exhibits exceptional reflection reliability. While existing internal reflection paradigms frequently suffer from performance degradation, our framework achieves consistent accuracy improvements of +2.82\% on LongVideoBench and +1.41\% on VideoMME. These results confirm that grounding self-correction in objective evidence effectively breaks the hallucination loop in long video understanding.

In summary, our main contributions are as follows:
\begin{itemize}
\item We propose Reflect-R1, the first Evidence-Driven self-correction framework. It effectively mitigates reflection failures caused by the lack of external evidence, enabling authentic self-correction grounded in objective clues.

\item We design SD-GRPO, a stage-decoupled reinforcement learning algorithm that independently computes advantage functions to overcome policy coupling in multi-stage reasoning. Additionally, we construct a dedicated dataset of 120K samples to bridge the training data gap for multimodal reflection.

\item Reflect-R1 achieves state-of-the-art performance on long video benchmarks, including VideoMME, LongVideoBench, and MLVU. Notably, it significantly improves the genuine rectification rate, demonstrating the high reliability of our proposed reflection paradigm.
\end{itemize}


    
        


\section{Can MLLM Correct Itself? A Preliminary Investigation}
\label{sec:Preliminary}

\begin{figure}[t]
  \centering
  \includegraphics[width=\linewidth]{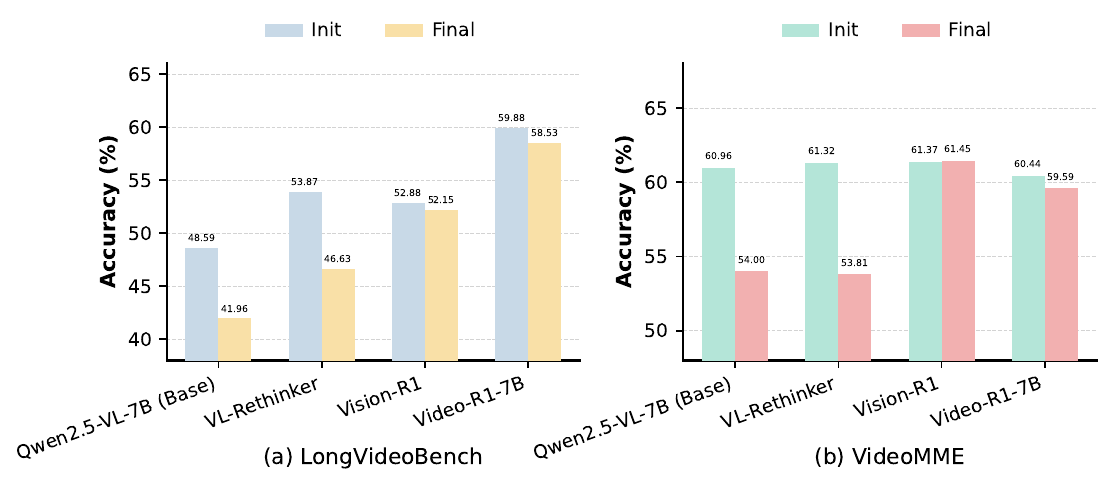}
    \caption{\textbf{The Failure of Internal Reflection.} Without objective external visual anchors, static closed-loop reflection tends to amplify initial visual hallucinations.}
  \label{fig:reliability_analysis}
\end{figure}

To investigate the self-correction capability of existing multimodal large language models in long video understanding, we conduct a preliminary empirical study on the LongVideoBench and VideoMME datasets. As shown in Fig.~\ref{fig:reliability_analysis}, we observe a counterintuitive phenomenon where relying solely on internal parametric knowledge for multi-turn reflection, transitioning from the initial intuition to the final response, fails to yield performance improvements and instead leads to a significant drop in accuracy. Specifically, the base model Qwen2.5-VL-7B experiences a sharp decline in accuracy on both benchmarks, dropping from 48.59\% to 41.96\% on LongVideoBench. Furthermore, even recent open-source models optimized specifically for multimodal reasoning, such as Video-R1-7B and Vision-R1, generally suffer from performance degradation during this purely internal reflection process.

To diagnose the root cause of this performance degradation, we analyze the behavioral logic of the models during the reflection process. In long video scenarios featuring extremely high visual information density and large temporal spans, the initial intuition of the model is highly susceptible to factual deviations due to missing keyframes or truncated contexts. Under these circumstances, forcing the model to self-correct without acquiring new visual evidence often causes it to over-rely on its initially generated erroneous context. This closed-loop reflection lacking external visual anchors prevents the model from establishing rigorous objective verification standards. Instead of accurately locating and correcting errors, the model tends to perpetuate or even amplify the initial visual hallucinations during repeated internal reasoning loops, ultimately leading to an overall accuracy decline where $\Delta$ Acc $< 0$.

The above analysis exposes the core limitation of closed-loop reflection where models lack independent and objective external visual evidence as a verification standard during the reasoning process. This insight directly motivates the core design of Reflect-R1. Relying strictly on internal knowledge prevents models from catching their own errors. To fix this, we must introduce external tools so the model can actively retrieve fresh evidence. Building upon this argument, we propose Reflect-R1 to achieve genuine self-correction by empowering the model with the capability to autonomously collect objective evidence and rigorously verify facts.

\section{Methodology}
\subsection{Problem Formulation and Inference Framework}
\label{sec:Problem Formulation}

Given a long video $V$ and a natural language question $q$, the objective is to generate an accurate textual response $y$. We model this procedure as a multi-step reasoning chain. Diverging from conventional approaches that directly approximate the single mapping distribution $P(y|V, q)$, we propose a dynamic decision-making process incorporating intuition, independent verification, and arbitration.

The core philosophy of Reflect-R1 is to train a single unified policy $\pi_{\theta}$ that internalizes intuition, verification, and arbitration behaviors, rather than training multiple independent sub-models. Operating within a multi-stage Markov Decision Process, the policy exhibits distinct reasoning behaviors conditioned on the current context state. 

In the intuition stage, conditioned on the raw video $V$ and question $q$, the policy leverages parametric intuition to rapidly generate an initial answer $y_1$ while autonomously invoking retrieval tools to localize a set of keyframes $F$.
\begin{equation}
    y_1, F \sim \pi_{\theta}(\cdot | V, q).
\end{equation}

\noindent During the verification stage, we enforce strict contextual isolation to ensure an independent evaluation. In this phase, the policy is denied access to the initial hypothesis $y_1$ and the global video $V$, with its input scope strictly restricted to a local context comprising only the question $q$ and the retrieved keyframes $F$. Relying exclusively on this retrieved evidence, the model generates an independent verification response $y_2$, thereby providing an unbiased assessment.

\begin{equation}
    y_2 \sim \pi_{\theta}(\cdot | F,q).
\end{equation}

\noindent In the arbitration stage, acting as the final arbitrator, the policy synthesizes the intuitive hypothesis $y_1$ and the independent verification $y_2$. To guarantee robustness, $\pi_{\theta}$ employs an active investigation mechanism where the model, regardless of consensus between $y_1$ and $y_2$, is mandated to re-invoke tools to backtrack through video $V$ for deep evidentiary re-confirmation, ultimately yielding the final answer $y_3$.

\begin{equation}
    y_3 \sim \pi_{\theta}(\cdot | V, F, q, y_1, y_2).
\end{equation}

While these behaviors are executed by a shared parameter set $\theta$, the task difficulty and reward scales vary significantly across stages. To address this, we propose the Stage-Decoupled GRPO algorithm, which fully decouples the advantage estimation for each stage during training. As detailed in Sec.~\ref{sec:method_algorithm}, this design prevents cross-stage competition during optimization. 

\subsection{The Dilemma of End-to-End Optimization} \label{sec:motivation}
\begin{figure}[t]
\centering
\includegraphics[width=\textwidth]{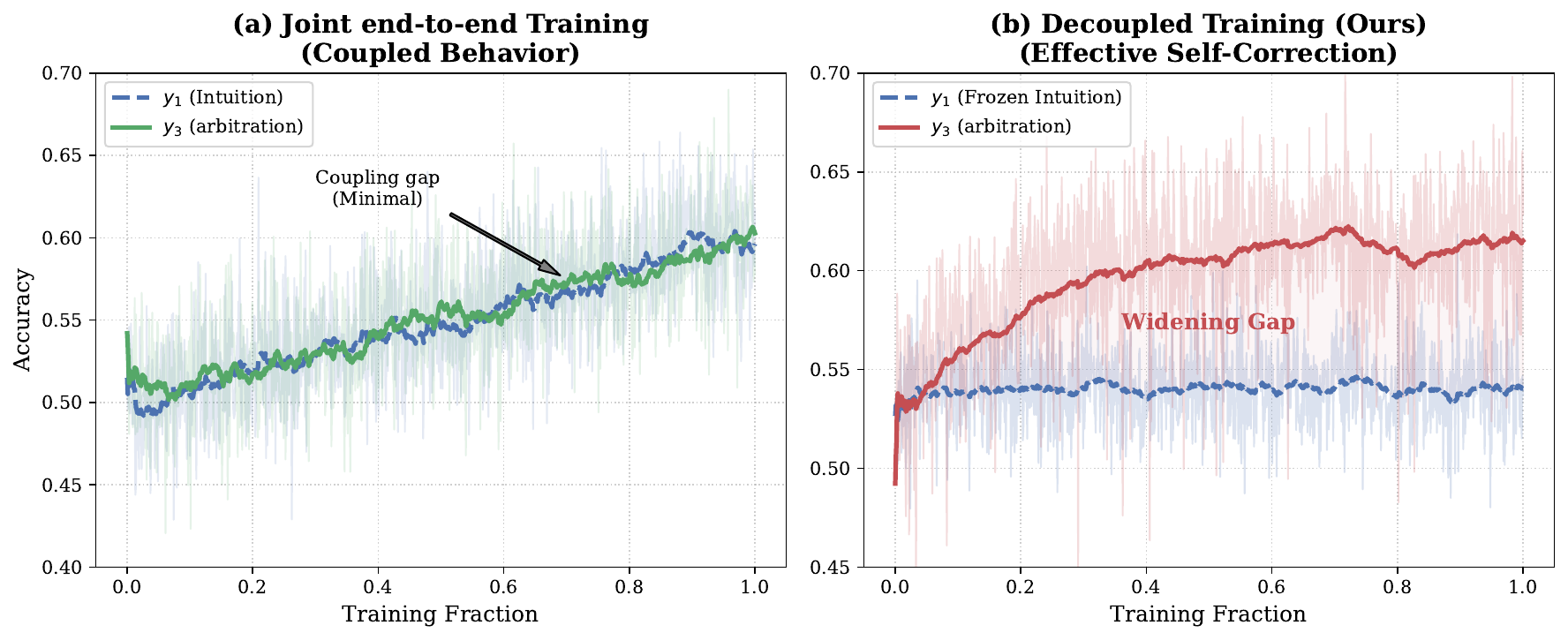}
\caption{\textbf{Decoupled training prevents policy coupling.} (a) Joint end-to-end training collapses reflection into a trivial identity mapping of the initial intuition. (b) Our decoupled strategy stabilizes the preceding distribution, enabling the model to learn robust error-correction logic and achieve a widening performance gap.}
\label{fig:pilot_study}
\end{figure}

Before detailing the stage-decoupled GRPO (SD-GRPO) algorithm, we analyze why a naive end-to-end optimization paradigm fails to elicit genuine self-correction capabilities in long video question answering. Specifically, the end-to-end approach attempts to simultaneously train intuition ($y_1$), verification ($y_2$), and arbitration ($y_3$) behaviors within a single training phase. As Figure \ref{fig:pilot_study} illustrates, we compare the training dynamics of this joint end-to-end strategy against our decoupled method.

Under the joint end-to-end training regime (Fig. \ref{fig:pilot_study} (a)), the model inevitably suffers from policy coupling. Because all reasoning stages undergo drastic gradient updates simultaneously, the arbitration policy $\pi_{\theta}$ tends to exploit optimization shortcuts. To rapidly secure base rewards, the policy directly copies the initial intuitive hypothesis $y_1$ instead of learning the complex logic required for error correction. This phenomenon eliminates the performance gap between $y_1$ and $y_3$, strips the model of its error-correction utility, and causes the arbitration behavior to degenerate into a trivial identity mapping.


In contrast, our decoupled training strategy (Fig.~\ref{fig:pilot_study} (b)) effectively resolves this issue through a two-stage design. By first stabilizing the intuition phase, we ensure that the subsequent arbitration stage learns to correct a stable set of initial errors. Empirical results demonstrate that the policy $\pi_{\theta}$ develops genuine error-correction capabilities only when the initial reasoning process remains stable, rather than simply memorizing the final answers. This finding directly confirms the necessity of the decoupled architecture in SD-GRPO, which relies on a stable foundation to unlock authentic self-reflection.

\begin{figure*}[t] 
    \centering
    \includegraphics[width=0.95\textwidth]{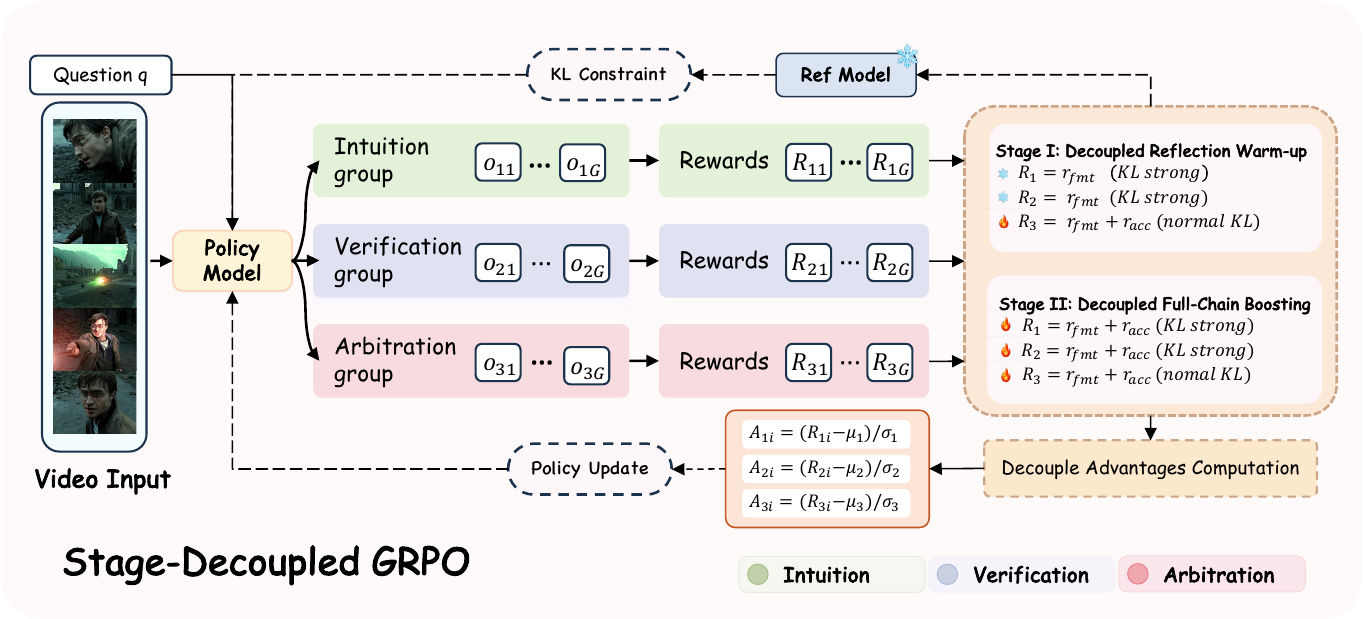} 
    \caption{\textbf{Overview of SD-GRPO.} We employ a progressive two-stage optimization: Stage I warms up the arbitration policy via strong KL constraints, and Stage II performs full-chain joint optimization. Additionally, group-wise advantage decoupling ensures that each reasoning stage evolves independently.}
    \label{fig:framework}
\end{figure*}

\subsection{Stage-Decoupled GRPO (SD-GRPO)}
\label{sec:method_algorithm}
Building upon the inference framework established in Sec.~\ref{sec:Problem Formulation}, we formalize the multi-step reasoning process as the sequential generation of three variables: intuition $y_1$, independent verification $y_2$, and arbitration $y_3$. To effectively optimize this long-chain reasoning process through GRPO~\cite{shao2024deepseekmath}, we propose the Stage-Decoupled GRPO (SD-GRPO) algorithm, as illustrated in Fig.~\ref{fig:framework}. This method addresses the credit assignment~\cite{arumugam2021information,nagpal2025leveraging,liu2019sequence} challenge inherent in varying reasoning depths through a group-wise advantage decoupling mechanism, while integrating stage-aware rewards and a progressive two-stage optimization to facilitate robust evolution from intuition to arbitration.

\noindent\textbf{Unified Objective and Advantage Decoupling.}
The overall optimization objective aims to maximize the cumulative expected return across the three reasoning stages. Formally, we define the total loss function $\mathcal{L}(\theta)$ as a weighted summation of the GRPO objectives corresponding to $y_1, y_2$, and $y_3$:

\begin{equation}
\label{eq:total_loss}
\begin{aligned}
\mathcal{L}(\theta) = \sum_{k=1}^{3} \mathbb{E}_{q \sim \mathcal{D}} \Bigg[ \frac{1}{G} \sum_{i=1}^{G} \frac{1}{L_{i,k}} \sum_{t=1}^{L_{i,k}} \bigg( & \mathcal{J}_{\text{clip}}^{(k)}(t, i) \\
& - \beta_k \mathbb{D}_{\text{KL}}\left( \pi_{\theta}(\cdot|x^{(k)}) \| \pi_{\text{ref}}(\cdot|x^{(k)}) \right)_t \bigg) \Bigg],
\end{aligned}
\end{equation}

\noindent where $\mathcal{J}_{\text{clip}}^{(k)}$ represents the PPO-based~\cite{pignatelli2024surveytemporalcreditassignment} clipped surrogate objective designed to stabilize policy updates:

\begin{equation}
\mathcal{J}_{\text{clip}}^{(k)}(t, i) = \min \left( \rho_{t,i}^{(k)} A_i^{(k)}, \text{clip}(\rho_{t,i}^{(k)}, 1-\epsilon, 1+\epsilon) A_i^{(k)} \right).
\end{equation}

\noindent Here, $\rho_{t,i}^{(k)} = \frac{\pi_{\theta}(y_{t,i}|x^{(k)})}{\pi_{\text{old}}(y_{t,i}|x^{(k)})}$ denotes the importance sampling ratio, and $A_i^{(k)}$ is the advantage term computed via group-wise normalization. The index $k \in \{1, 2, 3\}$ corresponds to the generation processes for $y_1, y_2$, and $y_3$ respectively. Accordingly, $x^{(k)}$ represents the stage-specific input context established in Sec.~\ref{sec:Problem Formulation}, where $x^{(1)} = \{V, q\}$, $x^{(2)} = \{F, q\}$, and $x^{(3)} = \{V, F, q, y_1, y_2\}$. The term $G$ represents the group size used for sampling, and $\beta_k$ controls the strength of the KL divergence penalty to prevent excessive deviation from the reference policy $\pi_{\text{ref}}$.

A critical challenge in multi-step reasoning is the significant disparity in task difficulty between the intuition ($y_1$) and arbitration ($y_3$) stages. Global normalization often causes the simpler intuition stage to dominate gradient updates because it inherently yields higher baseline rewards. To resolve this issue, we compute advantages independently within each generation stage. Specifically, for the $i$-th sample in the $k$-th stage, the advantage $A_{i}^{(k)}$ is defined as follows:

\begin{equation}
A_{i}^{(k)} = \frac{R_{i}^{(k)} - \mu_k}{\sigma_k + \epsilon},
\end{equation}

\noindent where $\mu_k$ and $\sigma_k$ are derived exclusively from the reward set $\{R_{1}^{(k)}, \dots, R_{G}^{(k)}\}$ associated with that specific stage. This design establishes a principle of intra-stage competition, where $y_1$ samples are compared solely against other $y_1$ samples, and similarly for $y_3$. This mechanism effectively isolates the reward distributions across different reasoning depths, ensuring that subtle improvement signals in the arbitration phase are not overshadowed by the high baselines inherent to the intuition phase.

\noindent\textbf{Stage-Aware Reward Function.}
To elicit differentiated reasoning behaviors across $y_1, y_2$, and $y_3$, we design a fine-grained reward function $R^{(k)} = r_{{fmt}} + r_{{acc}}^{(k)}$, where $r_{{fmt}}$ represents a universal format constraint reward and $r_{{acc}}^{(k)}$ is tailored to the specific characteristics of each stage.

The format reward $r_{{fmt}}$ serves as a structural regularizer to ensure syntactic correctness across all outputs. Specifically, it aggregates constraints from three dimensions: 1) Tag Adherence, which enforces the proper usage of XML delimiters; 2) Thought Length Reward, which encourages sufficient deliberation by regulating the length of the reasoning chain; and 3) Valid Tool Invocation, which verifies the syntactic accuracy and executability of API calls. Due to space constraints, the detailed mathematical formulations and implementation details of these format rewards are provided in the Appendix.

For the intuition stage ($k=1$), we employ a standard binary reward where $r_{\text{acc}}^{(1)} = \mathbb{I}(y_1 = y_{gt})$. In the verification stage ($k=2$), we introduce an honesty incentive to ensure the objectivity of the evaluation. Because $y_2$ is generated under a severely restricted field of view containing only the retrieved keyframes $F$, applying a standard binary reward would inevitably force the model into blind guessing when the provided visual evidence is insufficient. As recent studies on model abstention demonstrate~\cite{madhusudhan2025llms,tomani2024uncertainty,wei2025truthrl,godin2019learning}, forcing responses under partial information significantly exacerbates hallucinations. To address this, we design a ternary reward mechanism:


\begin{equation}
r_{\text{acc}}^{(2)} = 
\begin{cases} 
1, & \text{if } y_2 = y_{gt} \\ 
0, & \text{if } y_2 \in \mathcal{S}_{\text{abstain}} \\ 
-1, & \text{otherwise}
\end{cases},
\end{equation}

\noindent where $\mathcal{S}_{\text{abstain}}$ denotes the set of abstention responses (e.g.,~\texttt{"I don't know"}). This mechanism explicitly incentivizes the model to acknowledge ignorance when visual clues are inadequate, which yields a neutral score. It effectively penalizes errors resulting from baseless fabrication, thereby guaranteeing that the verification process remains strictly grounded in observable empirical evidence.

For the arbitration stage ($k=3$), we implement an anti-corruption penalty to prevent the model from overturning originally correct judgments. The reward is formulated as follows:

\begin{equation}
r_{\text{acc}}^{(3)} =
\begin{cases}
1, & \text{if } y_3 = y_{gt} \\
-1, & \text{if } y_3 \neq y_{gt} \land (y_1 = y_{gt} \lor y_2 = y_{gt}) \\
0, & \text{otherwise}
\end{cases}.
\end{equation}

\noindent This structure ensures that if the final answer $y_3$ is incorrect while at least one of the preceding outputs $y_1$ or $y_2$ is correct, the model incurs a strict penalty. This constraint compels the policy to exercise extreme caution and avoid destructive modifications during the final arbitration.

\begin{table}[t]
    \centering
    \footnotesize
    \caption{\textbf{Long video understanding performance.} We compare our method against state-of-the-art models. The baselines are categorized into two primary paradigms: standard inference relying solely on internal parameters (w/o Tools), and tool-augmented reasoning. $^\dagger$ indicates keyframes adaptively retrieved during the inference process.}
    \label{tab:video_understanding}
    \resizebox{1.0\textwidth}{!}{
        \begin{tabular}{l|c|cccc|c|c}
            \toprule
            \multirow{2}{*}{\textbf{Model}} & \multirow{2}{*}{\textbf{\# Frame}} & \multicolumn{4}{c|}{\textbf{VideoMME (w/o sub)}} & \textbf{MLVU} & \textbf{LVB} \\
            \cmidrule(lr){3-6} \cmidrule(lr){7-7} \cmidrule(lr){8-8}
            & & short & medium & long & \textbf{overall} & \textbf{m-avg} & \textbf{val} \\
            \midrule
            \rowcolor{gray!10}\multicolumn{8}{c}{\textit{w/o Tools}} \\
            Qwen2.5-VL-7B-Instruct~\cite{bai2025qwen25vltechnicalreport} & 768 & 71.4 & 60.1 & 52.3 & 61.3 & 57.9 & 53.9 \\
            GPT-4o~\cite{hurst2024gpt} & 384 & 80.0 & 70.3 & 65.3 & 71.9 & 64.6 & 66.7 \\
            Gemini-1.5-Pro~\cite{team2024gemini} & 1 fps & 81.7 & 74.3 & 67.4 & 75.0 & - & 64.0 \\
            
            \midrule
            VL-Rethinker~\cite{VL-Rethinker} & 768 & 70.1 & 60.8 & 53 & 61.3 & 63.5 & 55.58 \\
            Vision-R1-7B~\cite{huang2025vision} & 768 & 71.2 & 60.2 & 52.8 & 61.4 & 68.2 & 52.8 \\
            Video-R1-7B~\cite{feng2025video} & 768 & 72.2 & 58.1 & 52.3 & 60.8 & 68.5 & 60.1 \\
            \rowcolor{gray!10}\multicolumn{8}{c}{\textit{Tool-Augmented Reasoning}} \\
            VideoAgent (GPT-4)~\cite{wang2024videoagent} & 87$^\dagger$ & - & - & 49.0 & 56.0 & - & - \\
            T* (GPT-4o)~\cite{ye2025rethinking} & 32$^\dagger$ & 72.1 & 60.3 & 52.0 & 61.45 & - & - \\
            TimeSearch-R-7B~\cite{pan2025timesearch} & 768 & 73.5 & \textbf{61.2} & 53.4 & 62.7 & 69.3 & 61.2 \\
            \textbf{Reflect-R1 (Ours)} & 768 & \textbf{73.9} & {61.0} & \textbf{55.6} & \textbf{63.5} & \textbf{69.8} & \textbf{62.5} \\
            \bottomrule
        \end{tabular}
    }
\end{table}

\subsection{Training Strategies}

\subsubsection{Data construction.}
We aggregate video and question-answer pairs from six datasets: LLaVA-Video-178K~\cite{zhang2024llava}, Panda-70M~\cite{chen2024panda}, NExT-QA~\cite{xiao2021next}, PerceptionTest~\cite{patraucean2023perception}, CLEVRER~\cite{yi2019clevrer}, and STAR~\cite{wu2024star}. We employ Qwen2.5-VL-72B~\cite{bai2025qwen25vltechnicalreport} to synthesize chain-of-thought data aligned with our three-stage pipeline. A rule-based filter eliminates defective outputs to generate the Reflect-R1-CoT-90k dataset for SFT. Finally, GPT-4o~\cite{hurst2024gpt} selects 30,000 challenging samples to construct the Reflect-R1-RL-30k dataset for reinforcement learning. The appendix provides additional details.

\subsubsection{Model training.}
Reflect-R1 employs a two-stage training pipeline. First, supervised fine-tuning provides a cold start for the model to acquire structured reasoning formats and fundamental reflection paradigms. Subsequently, our SD-GRPO algorithm performs reinforcement learning post-training to unlock deep reasoning capabilities, enabling autonomous tool invocation and dynamic self-correction.


\section{Experiments}
\subsection{Setup}
\noindent\textbf{Benchmarks.}
We evaluate the proposed method on three widely adopted long-form video benchmarks: VideoMME~\cite{fu2025video}, LongVideoBench~\cite{wu2024longvideobench} and MLVU~\cite{zhou2025mlvu}.

\noindent\textbf{Training Details.}
We train the model using 8 NVIDIA H200 GPUs. During the training phase, we limit the maximum number of video frames to 734, processing each frame at a maximum resolution of $192 \times 28 \times 28$ pixels. During inference, we increase the frame capacity to 768 while maintaining all other configurations. For the reinforcement learning process, we set the group size $G$ to 8. More details are provided in Appendix.

\begin{table}[t]
    \centering
    \footnotesize
    \caption{\textbf{Temporal search performance.} We report temporal similarity, visual similarity, and question-answering (QA) accuracy on Haystack-LVBench. Baseline results are directly cited from~\cite{ye2025rethinking}. $^\dagger$ indicates the average number of keyframes determined by the model adaptively.}
    \label{tab:temporal_search}
    \resizebox{1.0\textwidth}{!}{
        \begin{tabular}{l|c|c|ccc|ccc|c}
            \toprule 
            \multirow{2}[2]{*}{\textbf{Method}} & \multirow{2}[2]{*}{\textbf{Base Model}} & \multirow{2}[2]{*}{\textbf{\# Frame}} & \multicolumn{3}{c|}{\textbf{Temporal}} & \multicolumn{3}{c|}{\textbf{Visual}} & \textbf{QA} \\
            \cmidrule(lr){4-6} \cmidrule(lr){7-9} 
            &  &  & P$\uparrow$ & R$\uparrow$ & $F_1$$\uparrow$ & P$\uparrow$ & R$\uparrow$ & $F_1$$\uparrow$ & LVBench \\
            \midrule
            \rowcolor{gray!10}\multicolumn{10}{c}{\textit{Static Frame Sampling}} \\ 
            Uniform  & Qwen2.5VL-7B & 8  & 1.4 & 6.3 & 2.2 & 56.0 & 72.0 & 62.7 & 33.7 \\
            Uniform  & GPT-4o & 8  & 1.4 & 6.3 & 2.2 & 56.0 & 72.0 & 62.7 & 47.1 \\
            \textcolor{gray}{Uniform} & \textcolor{gray}{GPT-4o} & \textcolor{gray}{32} & \textcolor{gray}{1.4} & \textcolor{gray}{24.9} & \textcolor{gray}{2.7} & \textcolor{gray}{58.7} & \textcolor{gray}{81.6} & \textcolor{gray}{67.3} & \textcolor{gray}{50.5}  \\
            \midrule
            \rowcolor{gray!10}\multicolumn{10}{c}{\textit{Adaptive Temporal Search}} \\
            VideoAgent~\cite{wang2024videoagent} & GPT-4  & 10.1$^\dagger$ & 1.2 & {8.5} & 2.1 & 58.8 & {73.2} & {64.7} & -  \\
            Retrieval-based~\cite{ye2025rethinking} & GPT-4o & 8  & 1.5 & 6.3 & 2.3 & {63.1} & 65.5 & 64.1 & -  \\
            T*~\cite{ye2025rethinking} & GPT-4o & 8 & {1.6} & 7.1 & {2.5} & 58.4 & 72.7 & 64.3 & {51.9} \\
            \textcolor{gray}{Retrieval-based~\cite{ye2025rethinking}} & \textcolor{gray}{GPT-4o} & \textcolor{gray}{32}  & \textcolor{gray}{1.3} & \textcolor{gray}{21.8} & \textcolor{gray}{2.4} & \textcolor{gray}{59.9} & \textcolor{gray}{80.8} & \textcolor{gray}{67.8}  & \textcolor{gray}{-}  \\
            \textcolor{gray}{T*~\cite{ye2025rethinking}} & \textcolor{gray}{GPT-4o} & \textcolor{gray}{32}  & \textcolor{gray}{1.7} & \textcolor{gray}{28.2} & \textcolor{gray}{3.1} & \textcolor{gray}{58.3} &  \textcolor{gray}{83.2} & \textcolor{gray}{67.8} & \textcolor{gray}{53.1} \\
            \midrule
            \rowcolor{gray!10}\multicolumn{10}{c}{\textit{Active Tool-Augmented Search}} \\
            TimeSearch-R~\cite{pan2025timesearch} & Qwen2.5VL-7B & 9.32$^\dagger$ & {5.5} &  {21.2} & {8.1} & {63.2} & \textbf{76.5} & {69.2} & {51.5} \\
            \textbf{Reflect-R1 (Ours)} & Qwen2.5VL-7B & 10.18$^\dagger$ & \textbf{6.3} &  \textbf{21.9} & \textbf{8.9} & \textbf{63.8} & {76.1} & \textbf{69.5} & \textbf{55.5} \\
            \bottomrule
        \end{tabular}
    }
\end{table}

\subsection{Main Results}

\noindent\textbf{Long-Form Video Understanding.} Reflect-R1 demonstrates remarkable performance across multiple challenging long-form video understanding benchmarks, as shown in Table~\ref{tab:video_understanding}. Reflect-R1 outperforms baseline models such as Qwen2.5-VL-7B by capturing definitive visual evidence through dynamic temporal search and performing reflective self-correction. Experimental results indicate that the performance margin of Reflect-R1 expands as the video duration increases, validating the robustness of its decoupled verification paradigm in processing complex long-temporal contexts.

\noindent\textbf{Temporal Search.} As shown in Table~\ref{tab:temporal_search}, Reflect-R1 outperforms state-of-the-art baselines in temporal similarity, visual similarity, and question-answering accuracy. This performance leap stems from our Evidence-Driven dynamic invocation mechanism. Unlike traditional single-pass retrieval pipelines, the arbitration stage proactively identifies verification failures caused by insufficient evidence and triggers tool re-invocation to retrieve the correct frames. Through the stage-decoupled optimization of SD-GRPO, the model learns an iterative and goal-oriented search strategy. This closed-loop feedback ensures that the retrieved visual evidence effectively supports complex reasoning, driving the synergistic evolution of keyframe localization precision and genuine self-correction capabilities.

\begin{table*}[t]
    \centering
    \caption{\textbf{Reflection Reliability Analysis.} We evaluate the self-correction capabilities across paradigms by comparing the initial intuitive accuracy and the final accuracy.}
    \label{tab:reflection_reliability}
    \resizebox{0.85\textwidth}{!}{ 
        \begin{tabular}{l|ccc|ccc}
            \toprule
            \multirow{2}{*}{\textbf{Model}} & \multicolumn{3}{c|}{\textbf{LongVideoBench}} & \multicolumn{3}{c}{\textbf{VideoMME}} \\
            \cmidrule(lr){2-4} \cmidrule(lr){5-7}
            & \textbf{Init} & \textbf{Final} & \textbf{$\Delta$ Acc} & \textbf{Init} & \textbf{Final} & \textbf{$\Delta$ Acc} \\
            \midrule
            \rowcolor{gray!10}\multicolumn{7}{c}{\textit{Internal Reflection (w/o Tools)}} \\
            Qwen2.5-VL-7B (Base)~\cite{bai2025qwen25vltechnicalreport} & 48.59\% & 41.96\% & \textcolor{red}{-6.63\%} & 60.96\% & 54.00\% & \textcolor{red}{-6.96\%} \\
            VL-Rethinker~\cite{VL-Rethinker} & 53.87\% & 46.63\% & \textcolor{red}{-7.24\%} & {61.32}\% & 53.81\% & \textcolor{red}{-7.51\%} \\
            Vision-R1~\cite{huang2025vision} & 52.88\% & 52.15\% & \textcolor{red}{-0.73\%} & \textbf{61.37}\% & {61.45\%} & \textcolor{teal}{+0.08\%} \\
            Video-R1-7B~\cite{feng2025video} & \textbf{59.88\%} & 58.53\% & \textcolor{red}{-1.35\%} & 60.44\% & 59.59\% & \textcolor{red}{-0.85\%} \\
            \midrule
            \rowcolor{gray!10}\multicolumn{7}{c}{\textit{Evidence-Driven Reflection}} \\
            \textbf{Reflect-R1 (Ours)} & {59.68\%} & \textbf{62.50\%} & \textbf{\textcolor{teal}{+2.82\%}} & {60.28\%} & \textbf{61.69\%} & \textbf{\textcolor{teal}{+1.41\%}} \\
            \bottomrule
        \end{tabular}
    }
\end{table*}

\noindent\textbf{Reflection Reliability.} As shown in Table~\ref{tab:reflection_reliability}, we evaluate self-correction efficacy by comparing initial and final accuracy. Baselines relying solely on internal parameters generally suffer from performance degradation ($\Delta$ Acc $< 0$). Without external visual anchors, closed-loop reflection amplifies visual hallucinations and frequently overturns initially correct judgments. In contrast, Reflect-R1 employs dynamic tool-use reflection to break this hallucination loop, achieving consistent accuracy improvements across benchmarks. This demonstrates that independent visual verification is crucial for overcoming reflection degradation in large vision-language models.

\subsection{Ablation Study}

\noindent\textbf{Component Ablation.} Tables~\ref{tab:ablation_strategy} and \ref{tab:ablation_reward} detail our ablation study on the LongVideoBench. Regarding training strategies, joint end-to-end optimization yields minimal performance gains from the intuition phase to final arbitration (from 61.2\% to 61.8\%), confirming that coupled training induces policy collapse. Bypassing the independent verification response $y_{2}$ while retaining active tool invocation fails to break the hallucination loop, causing the final accuracy to plummet to 55.9\%. Furthermore, replacing group-wise advantage decoupling with global calculation causes the reflection process to actually degrade arbitration accuracy below the model's own initial intuition (from 62.1\% down to 61.4\%), as simple intuitive tasks overshadow the complex optimization signals required for reflection. For reward formulation, ablating the abstention incentive forces the model to guess blindly under information bottlenecks, which prevents objective answer verification and misleads the final reflection, stalling the final accuracy at 60.8\%. Similarly, removing the anti-corruption penalty causes the model to frequently overturn initially correct judgments, resulting in an accuracy of 61.3\%. The complete Reflect-R1 framework integrates these stage-decoupled optimizations and fine-grained rewards to achieve the highest final accuracy of 62.50\%.

\begin{table}[t]
\centering

\newcommand{\CapBoxH}{3.2\baselineskip} 

\begin{minipage}[t]{0.49\linewidth}
\centering
\parbox[t][\CapBoxH][t]{\linewidth}{%
\captionof{table}{Training strategy ablation on LongVideoBench.}
\label{tab:ablation_strategy}%
}
\resizebox{\linewidth}{!}{
\begin{tabular}{l|ccc}
\toprule
\textbf{Strategy Variant} & \textbf{Init ($y_1$)} & \textbf{Verif ($y_2$)} & \textbf{Arbitration ($y_3$)} \\
\midrule
Joint end-to-end GRPO & 61.2 & 57.3 & 61.8 \\
w/o Verification ($y_2$) & 59.1 & - & 55.9 \\
w/o Advantage Decoupling & 62.1 & 57.4 & 61.4 \\
\midrule
\rowcolor{gray!10}\textbf{Reflect-R1 (Full)} & {59.68} & {56.62} & \textbf{62.50} \\
\bottomrule
\end{tabular}
}
\end{minipage}\hfill
\begin{minipage}[t]{0.49\linewidth}
\centering
\parbox[t][\CapBoxH][t]{\linewidth}{%
\captionof{table}{Reward function ablation on LongVideoBench.}
\label{tab:ablation_reward}%
}
\resizebox{\linewidth}{!}{
\begin{tabular}{l|ccc}
\toprule
\textbf{Reward Variant} & \textbf{Init ($y_1$)} & \textbf{Verif ($y_2$)} & \textbf{Arbitration ($y_3$)} \\
\midrule
w/o Abstention Incentive ($y_2$) & 59.2 & 61.2 & 60.8 \\
w/o Anti-Corruption Penalty ($y_3$) & 61.8 & 59.2 & 61.3 \\
\midrule
\rowcolor{gray!10}\textbf{Reflect-R1 (Full Rewards)} & {59.68} & {56.62} & \textbf{62.50} \\
\bottomrule
\end{tabular}
}
\end{minipage}
\end{table}

\begin{figure}[t] 
    \centering
    \includegraphics[width=\linewidth]{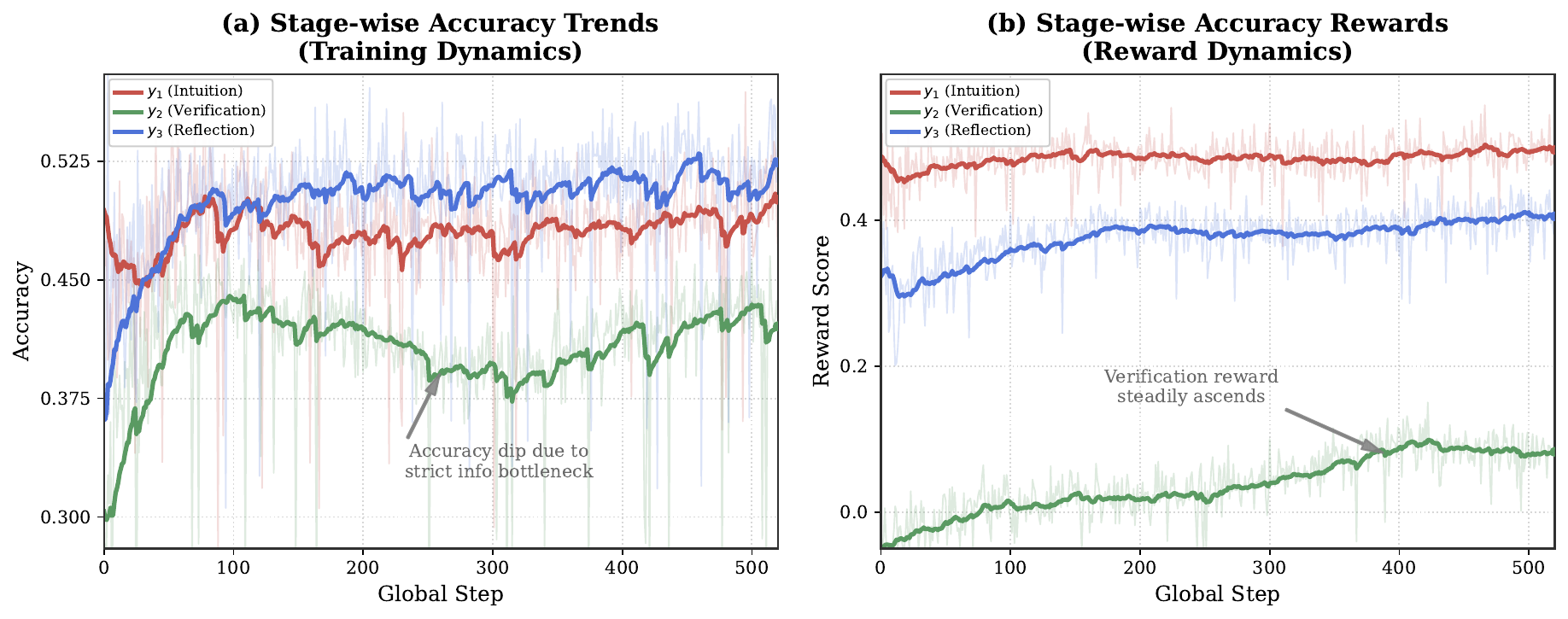}
    \caption{\textbf{Training Dynamics of Stage-Decoupled Verification.}}
    \label{fig:training_dynamics}
\end{figure}

\begin{figure*}[th]
    \centering
    \includegraphics[width=1.0\textwidth]{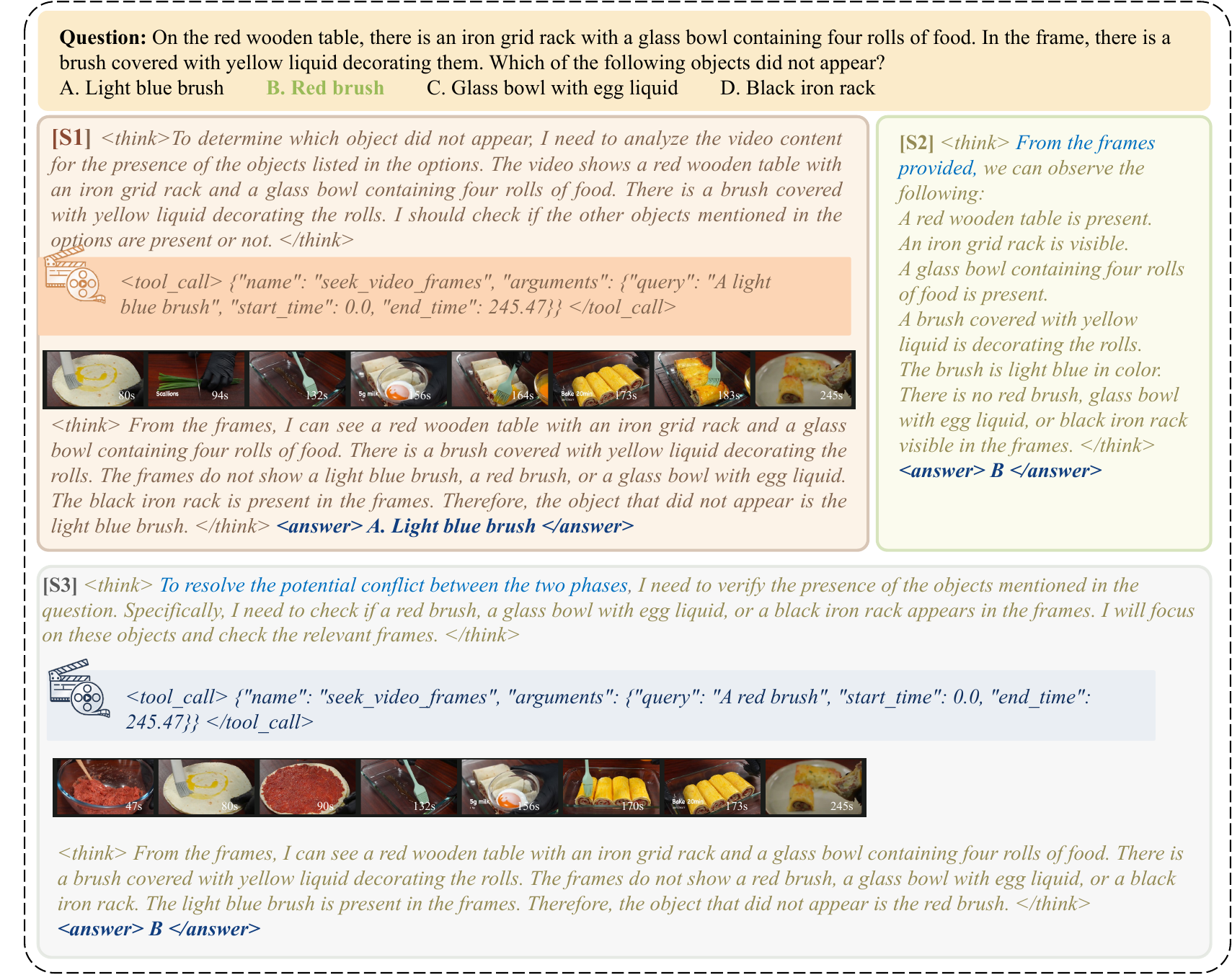}
    \caption{\textbf{Qualitative example of the reflective reasoning process.}}
    \label{fig:visualization}
\end{figure*}

\noindent\textbf{Analysis of Training Dynamics.} Fig.~\ref{fig:training_dynamics} illustrates the training trajectory over the first 500 steps. The distinct stratification of the reward curves (b) demonstrates that SD-GRPO successfully decouples the reasoning stages and effectively prevents policy coupling. The accuracy of the blind-test verification ($y_2$) experiences an initial decline, yet its corresponding reward climbs steadily from a negative value (a). This phenomenon validates the effectiveness of our reward design. Constrained by the information bottleneck, the model learns to abandon speculative guessing and instead formulates answers strictly based on visual evidence. Furthermore, despite the decoupled policies, the demand of $y_2$ for high-quality evidence acts as environmental feedback. This feedback compels the intuitive stage ($y_1$) to continuously improve its retrieval precision, which drives a synergistic evolution between the two stages. Ultimately, the reflective arbitration ($y_3$) utilizes this reliable evidence to correct initial errors and establishes a widening performance gap with $y_1$. This dynamic conclusively proves the emergence of genuine self-correction capabilities within the model.

\section{Qualitative Analysis}
Figure~\ref{fig:visualization} illustrates the autonomous self-correction of Reflect-R1. When the initial intuition ($y_1$) suffers from visual hallucinations, the independent verification ($y_2$) provides objective counter-evidence. The reflective arbitration ($y_3$) then resolves this conflict through adaptive temporal search to derive the correct answer. This process confirms that SD-GRPO effectively breaks hallucination loops via evidence-based reasoning.

\section{Limitations}
Although Reflect-R1 performs well on long-video reasoning, it has several limitations. First, the three-stage procedure often requires multiple tool calls and long generations, which increases inference latency and compute cost, so it is less suitable for strict real-time settings. Second, the method relies on temporal retrieval to obtain evidence. When retrieval is inaccurate or the key cues are weak, the subsequent verification and arbitration stages may have insufficient support to correct errors. Finally, overall performance still depends on the underlying vision-language model, which can fail on a small number of cases with high complexity or ambiguous evidence. Future work focuses on improving inference efficiency and integrating a broader set of more accurate tools to strengthen evidence acquisition and robustness.


\section{Conclusion}
In this work, we propose Reflect-R1, an Evidence-Driven reflection framework integrating stage-decoupled verification and dynamic tool invocation to address the self-correction challenge in long video understanding. To overcome the Internal Closed-Loop Reflection and the policy coupling in multi-stage reinforcement learning, we design the SD-GRPO algorithm. This algorithm drives the synergistic evolution of intuition, verification, and arbitration through intra-group advantage isolation. Reflect-R1 achieves state-of-the-art performance on benchmarks including VideoMME, LongVideoBench and MLVU, enabling genuine self-correction strictly grounded in objective evidence. We hope this work makes a meaningful contribution to advancing reinforcement learning for reflection in multimodal large language models.

\clearpage
\section*{Acknowledgements}
\begin{sloppypar}
This work was supported in part by the Guangdong Basic and Applied Basic Research Foundation (2026A1515010184), the Special Foundations for the Development of Strategic Emerging Industries of Shenzhen (No. KJZD\allowbreak20231023094700001), and the Shenzhen-Tsinghua Special Project for Fundamental \& Frontier Research in Artificial Intelligence (No. AI\allowbreak2026018).
\end{sloppypar}
\clearpage



%
%
\bibliographystyle{splncs04}
\bibliography{main}
\clearpage
\appendix
\renewcommand{\thesection}{\Alph{section}}
\setcounter{section}{0}
\graphicspath{{Figures/}{Supplementary_Material/}}
\newcommand{\ReflectRSkipSupplementReferences}{}
\ifdefined\ReflectRSkipSupplementReferences
\else
\PassOptionsToPackage{table}{xcolor}
\documentclass[runningheads]{llncs}

 
\usepackage{eccv}



\usepackage{eccvabbrv}
\renewcommand{\thesection}{\Alph{section}}
\usepackage{graphicx}
\usepackage{booktabs}
\usepackage{booktabs}
\usepackage{multirow}
\usepackage{graphicx}
\usepackage{xcolor}
\usepackage{pifont}
\usepackage[table]{xcolor}
\definecolor{gray}{gray}{0.4}

\usepackage[accsupp]{axessibility}  
\usepackage{placeins}


%

\usepackage{hyperref}

\usepackage{orcidlink}

\setlength{\textfloatsep}{6pt plus 2pt minus 2pt}
\setlength{\floatsep}{6pt plus 2pt minus 2pt}
\setlength{\intextsep}{6pt plus 2pt minus 2pt}
\setlength{\dbltextfloatsep}{6pt plus 2pt minus 2pt}
\setlength{\dblfloatsep}{6pt plus 2pt minus 2pt}
\setlength{\abovecaptionskip}{3pt plus 1pt minus 1pt}
\setlength{\belowcaptionskip}{0pt plus 1pt minus 1pt}
\captionsetup{skip=3pt}

\begin{document}
\fi

\begin{center}
    {\Large\bfseries Reflect-R1: Evidence-Driven Reflection for Self-Correction in Long Video Understanding}\\[0.5em]
    {\large\bfseries Supplementary Material}
\end{center}

\noindent This appendix provides additional details of our method. It is organized as follows:

\noindent Section~\ref{A}: Related Work\\
Section~\ref{B}: Reward Details\\
Section~\ref{C}: Dataset Construction Details\\
Section~\ref{D}: Training Details\\
Section~\ref{E}: Prompt Design\\
Section~\ref{F}: More Case Studies

\section{Related Work}
\label{A}
Reflective reasoning has emerged as a prominent research direction to enhance the reliability of large language models. Early efforts predominantly focus on self-feedback and iterative refinement within the pure text domain. For example, Self-Refine~\cite{madaan2023self} generates feedback from the same model to iteratively rewrite the output, whereas Reflexion~\cite{shinn2023reflexion} utilizes reflective memory in natural language to improve subsequent sequential decisions. These approaches demonstrate that models can significantly improve output quality through self-examination during the reasoning phase. However, their core mechanisms remain constrained by closed-loop modifications based on internal representations, lacking explicit reliance on external objective facts. ReAct~\cite{yao2022react} further intertwines reasoning and acting, enabling models to invoke external tools or knowledge sources during the thought process. This establishes the foundation for subsequent agentic reasoning and tool-augmented mechanisms.

In the field of multimodal reasoning, recent studies begin to integrate reflection mechanisms and reinforcement learning into vision-language models. Works such as Vision-R1~\cite{huang2025vision} and Video-R1~\cite{feng2025video} explore how to directly elicit the long-chain reasoning and alignment capabilities of large multimodal models through vision-guided reinforcement learning. Furthermore, VL-Rethinker~\cite{VL-Rethinker} explicitly encourages self-reflection in models using reinforcement learning. These studies demonstrate the substantial potential of reflection mechanisms and reinforcement learning in multimodal scenarios. Nonetheless, they primarily target static images or general visual question answering. They fail to systematically address the coordination challenges of evidence retrieval, independent verification, and final arbitration caused by massive information streams in long video contexts. In contrast, we focus on evidence-driven self-correction in long video understanding. We extend the reflection paradigm from a closed-loop review completely dependent on internal parameters to a rigorous verification process grounded in external temporal objective evidence.

Another research trajectory closely related to our work involves tool-augmented reasoning, structured representation, and active temporal retrieval in long video understanding. VideoAgent~\cite{wang2024videoagent} treats the large model as a central agent to collect key information in videos through iterative planning and tool invocation. From a representation perspective, VideoTree~\cite{wang2025videotree} proposes a query-adaptive hierarchical tree structure to compress redundant information in a coarse-to-fine manner, thereby supporting efficient long-temporal reasoning. Additionally, the representative work by Ye et al.~\cite{ye2025rethinking} points out that the core difficulty of long video understanding lies in accurately locating a minimal amount of key evidence from massive frames, and accordingly proposes the T* temporal retrieval framework. Concurrently, TimeSearch-R~\cite{pan2025timesearch} combines adaptive temporal retrieval with a self-verification mechanism, emphasizing the decisive role of retrieval quality in subsequent reasoning. These methods collectively indicate that relying solely on static inputs or one-time sampling usually encounters performance bottlenecks when processing long videos. Models must possess the capability to actively search, compress, and organize relevant segments. Our work aligns with them in emphasizing the importance of external retrieval. Advancing this concept, we propose a three-stage framework comprising intuition, verification, and arbitration. By explicitly decoupling the optimization objectives of different cognitive stages through SD-GRPO, we  unify tool invocation, objective evidence verification, and final noise-resistant arbitration within a single evidence-driven reflection framework.

\section{Reward Details}
\label{B}
\subsection{Format Reward Details}
As introduced in Section 3.3, we incorporate a format reward ($r_{fmt}$) to constrain the behavior of the policy model during multi-stage reasoning. This reward ensures structural consistency in the outputs and prevents reward hacking, a prevalent issue in reinforcement learning. The format reward linearly combines three fine-grained constraint components: tag adherence, thought length reward, and valid tool invocation reward.

\noindent\textbf{Tag Adherence.} We impose strict XML format constraints on the model outputs. The responses must include explicit reasoning tags, specifically \textless think\textgreater\ and \textless/think\textgreater. These must be followed by either answer tags (\textless answer\textgreater\ and \textless/answer\textgreater) or tool invocation tags (\textless tool\_call\textgreater\ and \textless/tool\_call\textgreater). We formulate a rigorous binary format detection function $r_{\mathrm{tag}}$ using indicator variables for correct tag closures. Let $\mathcal{I}_{\mathrm{think}}$, $\mathcal{I}_{\mathrm{ans}}$, and $\mathcal{I}_{\mathrm{tool}}$ represent the successful parsing of their respective tag pairs. The reward is defined as:

$$r_{\mathrm{tag}} =
\begin{cases}
1.0, & \text{if } \mathcal{I}_{\mathrm{think}} = 1 \text{ and } (\mathcal{I}_{\mathrm{ans}} \oplus \mathcal{I}_{\mathrm{tool}}) = 1, \\[4pt]
0.0, & \text{otherwise}.
\end{cases}$$

\noindent where $\oplus$ denotes the logical exclusive OR operation. If the model mixes tool invocations and final answers in the same turn or exhibits tag closure errors, the sequence is immediately truncated and receives the zero reward.

\noindent\textbf{Thought Length Reward.}
In long-chain reasoning, excessively short thoughts often indicate that the model is taking optimization shortcuts to guess blindly. Conversely, unconstrained and verbose reasoning inevitably introduces hallucinations and wastes computational resources. To address this, we design a piecewise length reward mechanism featuring soft boundaries and an asymmetric penalty. Let $l$ represent the length of the text generated within the \textless think\textgreater\ tags, while $L_{min}$ and $L_{max}$ denote the lower and upper bounds of the optimal reasoning length. In our experiments, these bounds are set to 120 and 700 respectively. The length reward $r_{len}$ is formulated as follows:

\[
r_{\mathrm{len}} =
\begin{cases}
l / L_{\min}, & \text{if } 0 < l < L_{\min}, \\[4pt]
1.0, & \text{if } L_{\min} \le l \le L_{\max}, \\[4pt]
1.0 - \dfrac{l - L_{\max}}{L_{\max}}, & \text{if } L_{\max} < l \le 2L_{\max}, \\[8pt]
-1.0, & \text{if } l > 2L_{\max}.
\end{cases}
\]

\noindent\textbf{Valid Tool Invocation Reward.} We design a tool invocation incentive specifically for the intuition and arbitration stages, as both phases require the active acquisition of objective evidence. To prevent reward hacking where the model exploits system fallback parameters to accumulate tool-use scores, we introduce posterior verification based on execution results. Let $a_{\mathrm{tool}}$ denote the parsed arguments of the tool call and $E(a_{\mathrm{tool}})$ represent the execution feedback indicating the number of successfully retrieved frames. The valid tool invocation reward $r_{\mathrm{tool}}$ follows a strict binary formulation:

$$r_{\mathrm{tool}} =
\begin{cases}
1.0, & \text{if parsing succeeds and } E(a_{\mathrm{tool}}) > 0, \\[4pt]
0.0, & \text{otherwise}.
\end{cases}$$

\noindent This ensures that a tool invocation is deemed valid only when it returns actual visual information.

\subsection{The Necessity of the Abstention Incentive in Verification}

\begin{figure}[!h]
\centering
\includegraphics[width=\linewidth]{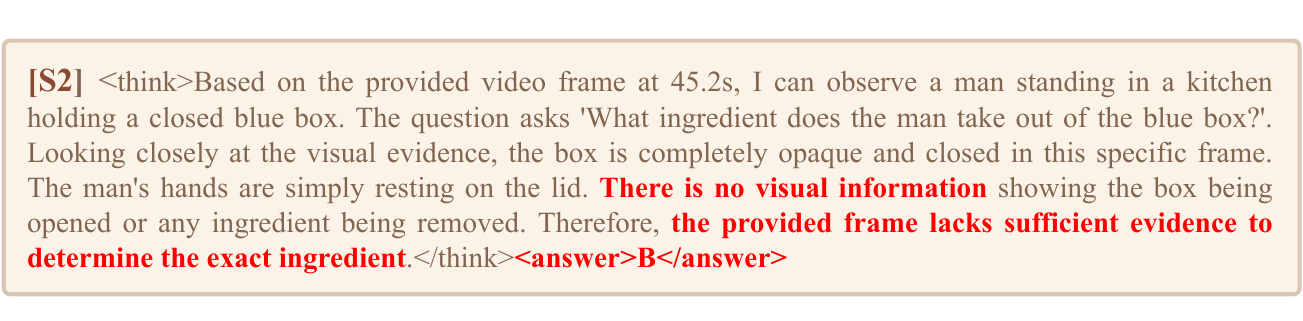}
\caption{A disconnect between reasoning and response occurs during the blind verification stage (S2) due to the absence of an abstention incentive. The model accurately identifies the lack of visual evidence during its reasoning process but is forced to make an ungrounded guess in the final output.}
\label{fig:abstention_badcase}
\end{figure}

Section 3.3 introduces a ternary reward mechanism with an abstention incentive for the independent verification stage. We illustrate the necessity of this design through a diagnostic failure case in Fig. \ref{fig:abstention_badcase}. Under a standard binary reward formulation, forcing the model to guess severely impairs the training process. If a baseless guess happens to match the ground truth, the model receives a reward. This feedback reinforces lucky guessing behaviors and prevents the model from conducting objective verification based on the available visual evidence. Consequently, the model generates misleading answers that interfere with the third-stage arbitration. The ternary abstention reward eliminates this issue by allowing the model to abstain and receive a neutral reward when evidence is insufficient. This mechanism forces the verification stage to act as a rigorous evidence reviewer. More importantly, an honest abstention explicitly triggers the arbitration stage to reinvoke temporal search tools. By actively mining new visual clues from the global video, the framework successfully completes a genuine closed-loop verification.

\section{Dataset Construction Details}
\label{C}
\begin{figure}[h]
\centering
\includegraphics[width=1\textwidth]{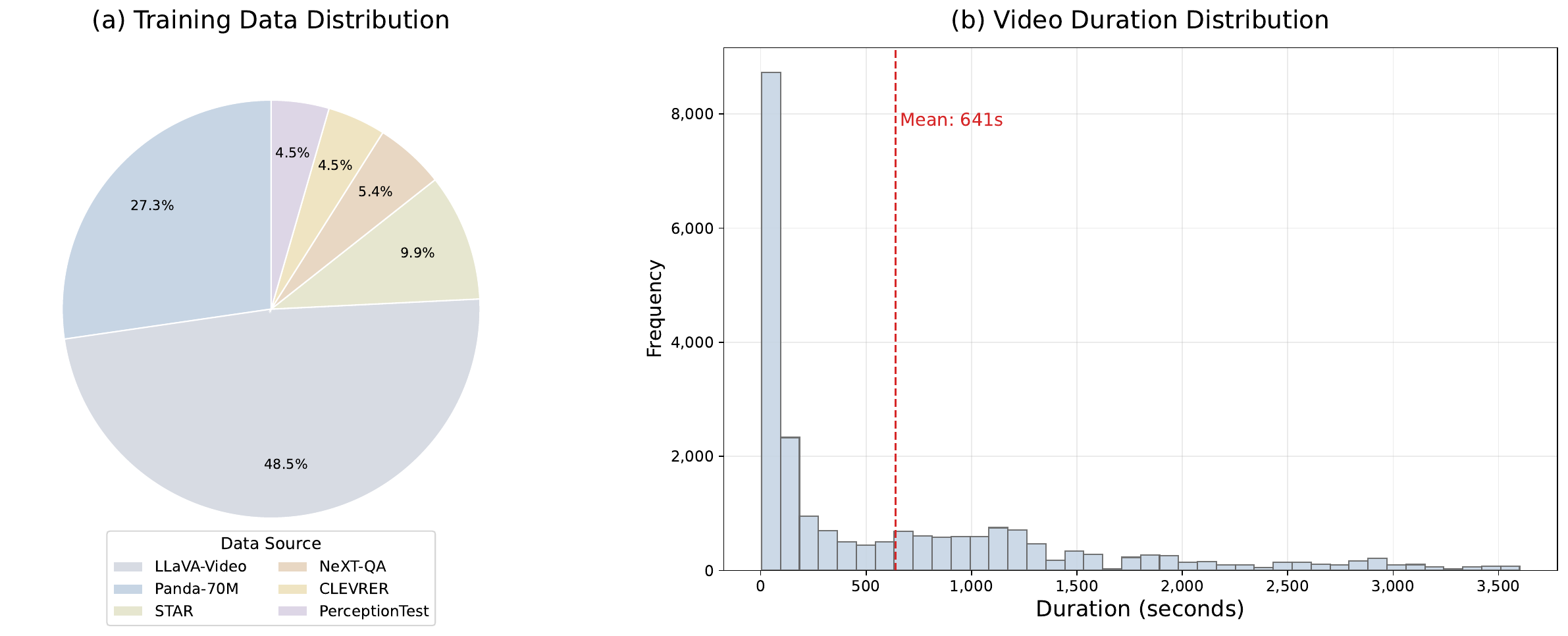}
\caption{Training data distribution across different source datasets and video durations.}
\label{fig:data_distribution}
\end{figure}

Fig.~\ref{fig:data_distribution} illustrates the characteristics of the training data. The data sources primarily consist of LLaVA-Video (48.5\%) and Panda-70M (27.3\%), supplemented by STAR, NeXT-QA, CLEVRER, and PerceptionTest. Additionally, the video duration distribution spans a broad range with a mean of 641 seconds, which provides a solid foundation for evaluating long-context reasoning capabilities. Qwen2.5-VL-72B is used to generate the supervised fine-tuning data. For each question, the model concurrently produces responses for the intuition, verification, and arbitration stages. An initial filtering step is then applied based on the correctness of these generated responses. The intuition and arbitration subsets are further refined by retaining only samples that contain valid tool invocations. A balanced proportion across the three stages is maintained throughout the construction process. For reinforcement learning data selection, GPT-4o receives four frames sampled from a long video to answer the corresponding question. Questions that can be answered correctly under this limited visual context are regarded as invalid and the associated long-video samples are discarded.

\section{Training Details}
\label{D}
We employ a cosine learning rate decay schedule for both the supervised fine-tuning cold start and the reinforcement learning phases. The model undergoes supervised fine-tuning for a single epoch. Subsequently, we conduct the first stage of reinforcement learning to stabilize the reflection capabilities and fundamental output formatting. This is followed by the second stage of reinforcement learning. During the optimization process, the Kullback-Leibler divergence coefficients $\beta_1$, $\beta_2$, and $\beta_3$ for the three reasoning stages are set to 0.05, 0.05, and 0.005, respectively.

\section{Prompt Design}
\label{E}
This section details the specific prompts utilized across the system level and the three distinct reasoning stages. These prompts rigorously regulate the output formatting, tool invocation protocols, and the abstention mechanism of the model. This section also includes the reflection prompt used by the internal closed-loop reflection baseline for comparison.

\begin{figure}[htbp]
\centering
\includegraphics[width=0.9\linewidth]{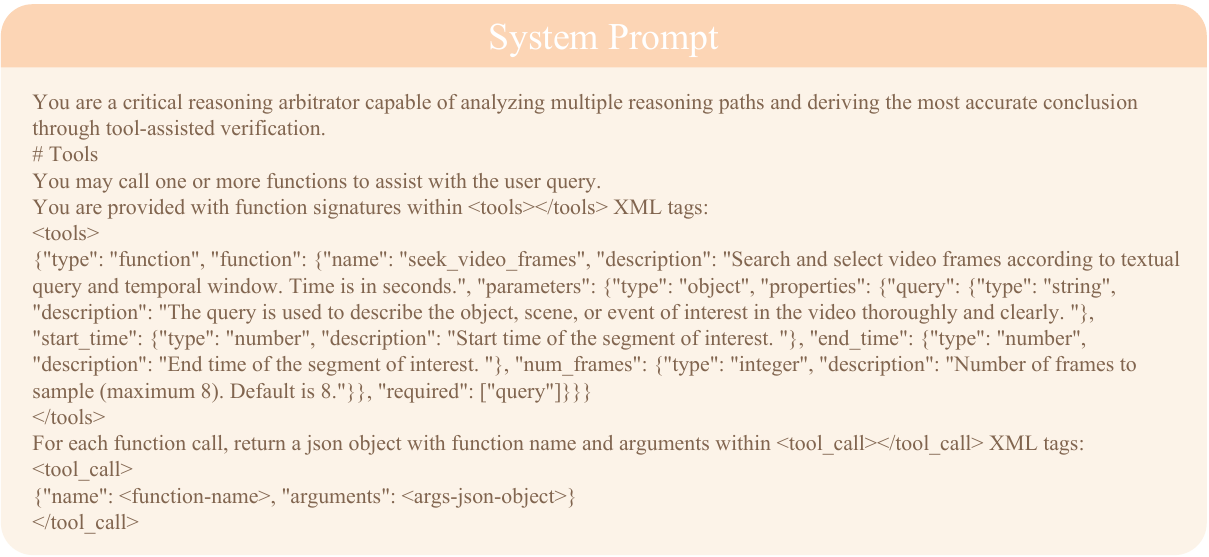}
\caption{The system prompt used to define the model persona and tool signatures.}
\label{fig:prompt_system}
\end{figure}

\noindent\textbf{System Prompt.} We follow the official tool-use protocol of the Qwen2.5-VL series and adopt its \textless tool\_call\textgreater~format to invoke the temporal retrieval tool, as shown in Fig.~\ref{fig:prompt_system}. This design standardizes the instruction format for tool invocation, allowing the model to generate executable retrieval commands and thereby obtain the video frames required for arbitration with precision.

\begin{figure}[htbp]
\centering
\includegraphics[width=0.9\linewidth]{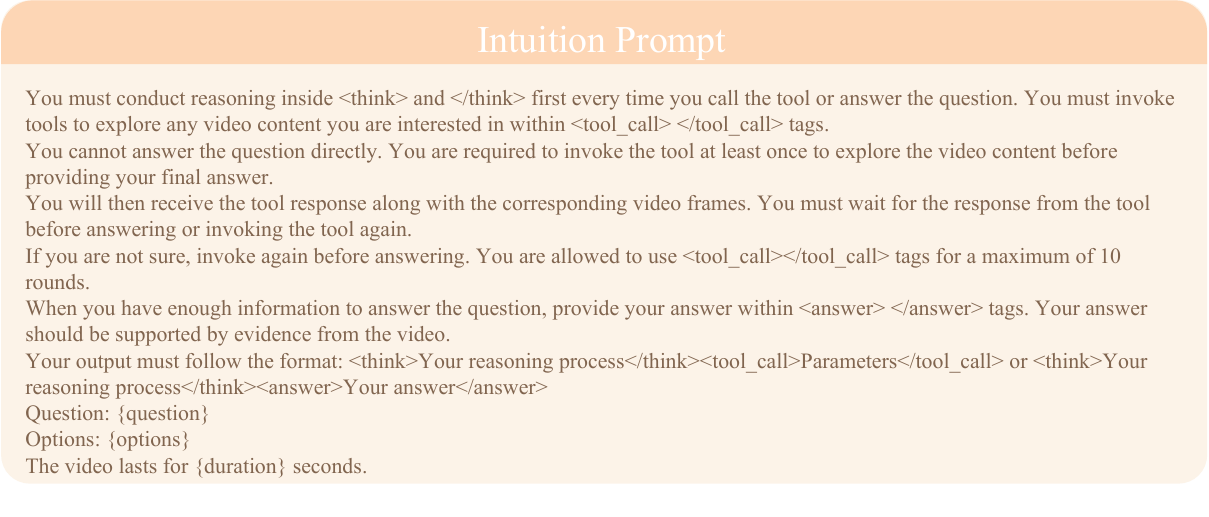}
\caption{The intuition prompt (S1) requiring active tool exploration.}
\label{fig:prompt_s1}
\end{figure}

\noindent\textbf{Intuition Prompt.} During the first stage, the prompt mandates that the model must invoke the retrieval tool at least once to explore the video content before providing a final answer, as detailed in Fig. \ref{fig:prompt_s1}. The policy is permitted a maximum of 10 interaction rounds, and it is strictly required to conduct chain-of-thought reasoning within designated tags prior to executing any action.

\begin{figure}[h]
\centering
\includegraphics[width=0.9\linewidth]{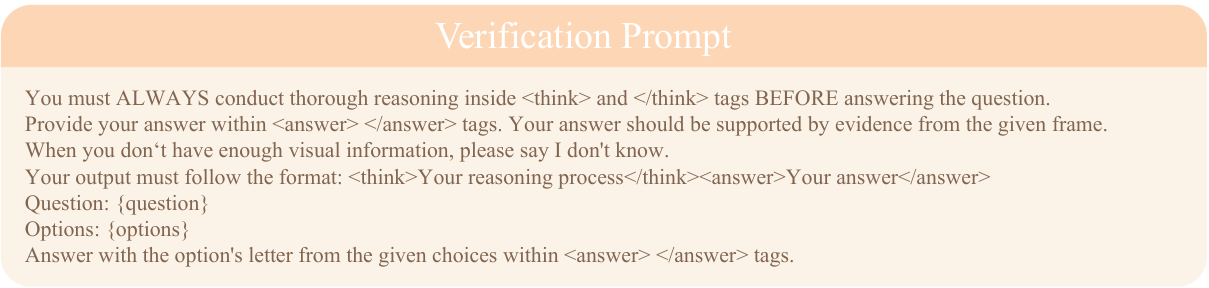}
\caption{The verification prompt (S2) incorporating the abstention mechanism.}
\label{fig:prompt_s2}
\end{figure}

\noindent\textbf{Verification Prompt.} The second-stage prompt severely restricts the context, compelling the model to base its reasoning exclusively on the provided local image frames (see Fig. \ref{fig:prompt_s2}). Crucially, this prompt integrates the abstention mechanism by explicitly instructing the model to output an acknowledgment of ignorance when visual information is insufficient. This instruction effectively prevents forced hallucination under information bottlenecks.

\begin{figure}[h]
\centering
\includegraphics[width=0.9\linewidth]{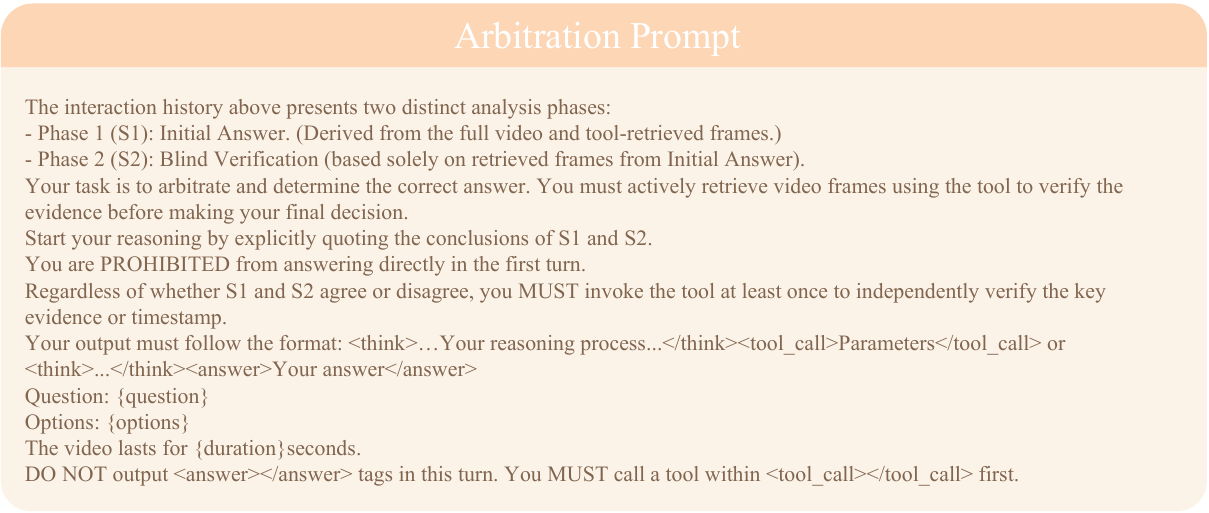}
\caption{The arbitration prompt (S3) designed for conflict resolution.}
\label{fig:prompt_s3}
\end{figure}

\noindent\textbf{Arbitration Prompt.} The prompt for the third stage presents the model with the complete interaction history of the preceding phases, as shown in Fig. \ref{fig:prompt_s3}. It forces the policy to explicitly extract and quote the conclusions from the first two stages at the beginning of its reasoning process. Furthermore, it strictly prohibits direct answering in the initial turn, compelling the model to autonomously invoke tools to verify key evidence independently, which facilitates genuine self-correction.

\begin{figure}[h]
\centering
\includegraphics[width=0.9\linewidth]{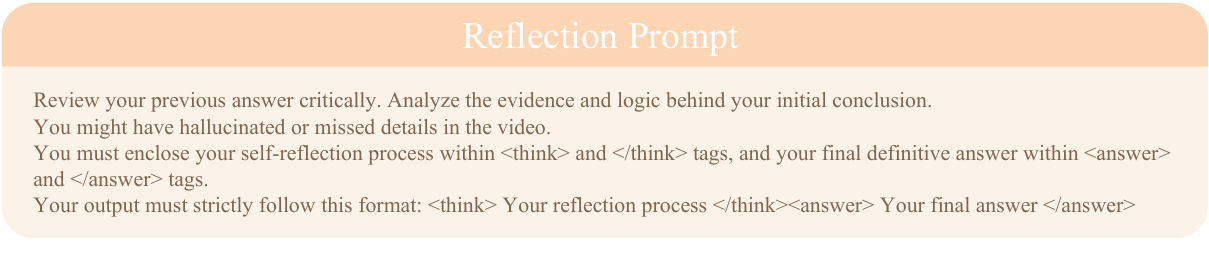}
\caption{Reflection prompt used by the internal closed-loop reflection baseline.}
\label{fig:prompt_reflect}
\end{figure}

\noindent\textbf{Reflection Prompt (Baseline).}
For comparison, we also construct an internal closed-loop reflection baseline that revisits the model's previous answer using a dedicated reflection prompt. This prompt instructs the model to critically review the reasoning behind its
initial decision and revise the answer if necessary without accessing any
additional visual evidence. The exact prompt used in our experiments is shown
in Fig.~\ref{fig:prompt_reflect}.

\section{More Case Studies}
\label{F}
To further illustrate the autonomous self-correction capabilities elicited by SD-GRPO, we present additional qualitative cases in this section.

\subsection{Failure Modes of Internal Closed-Loop Reflection}
\begin{figure}[t]
\centering
\includegraphics[width=1\textwidth]{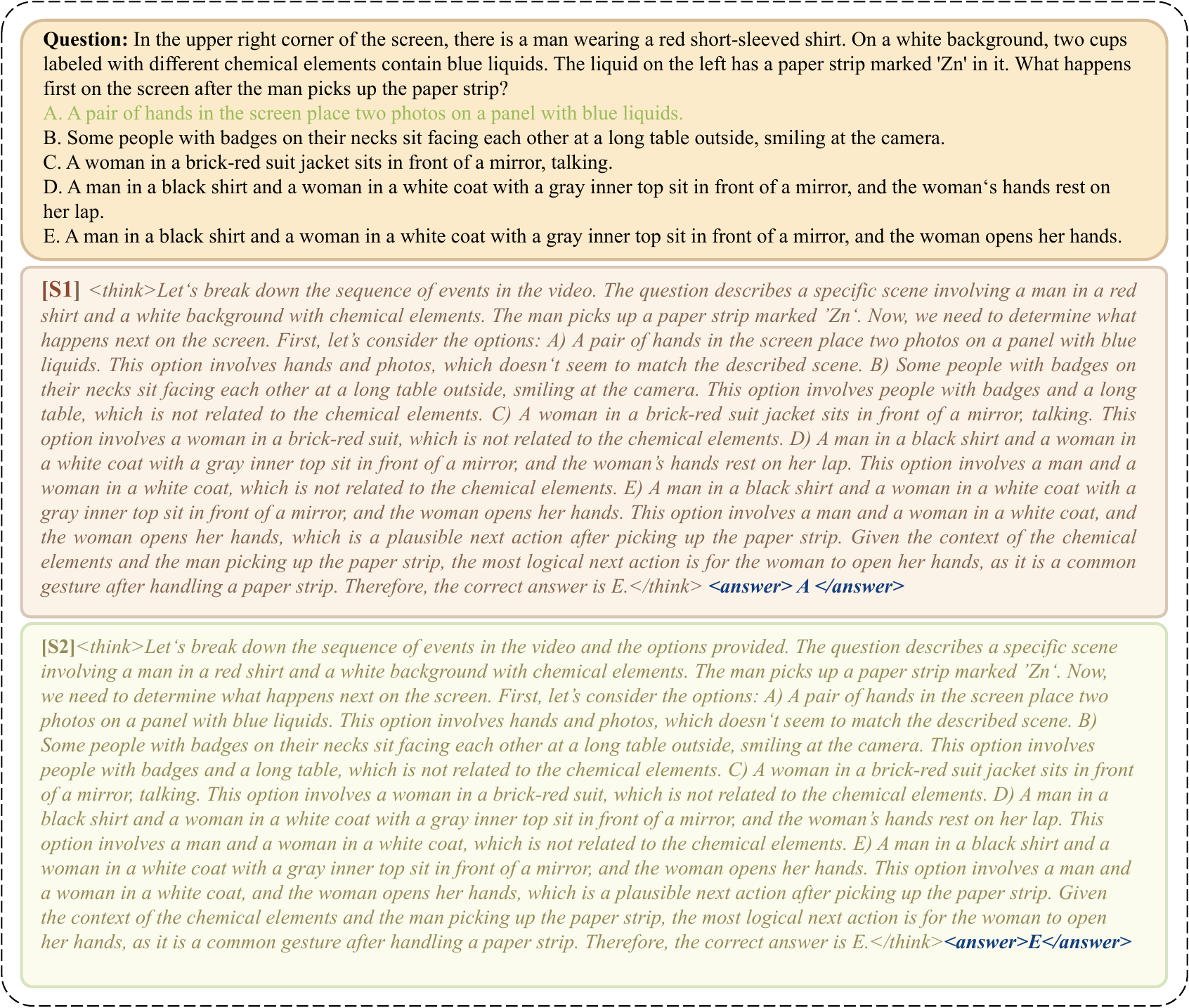}
\caption{
\textbf{Overturning a Correct Initial Answer during Internal Reflection.} Without external objective anchors, the model becomes misled by internally generated hallucinations during reflection, mistakenly overturning its initially correct answer.
}
\label{fig:internal_fail_2}
\end{figure}

We first present the failure modes of internal closed-loop reflection to further clarify the empirical observations discussed in Section~2. When objective visual evidence is absent, models that rely entirely on internal knowledge typically exhibit two dominant failure trajectories. The first trajectory manifests as the mechanical repetition of an initially incorrect prediction (see Fig.~\ref{fig:internal_fail_1}). In this case, the reflection stage does not introduce new insights but instead reproduces the flawed reasoning path contained in the initial intuition. As a result, the model enters a self-reinforcing loop that consolidates the original visual hallucination. The second trajectory manifests as the erroneous overturning of a correct judgment (see Fig.~\ref{fig:internal_fail_2}). When the model processes complex details in long videos without external visual anchors, its internal reasoning often becomes unstable. Even when the model produces a correct answer in the initial stage, the subsequent reflection process can be misled by internally generated hallucinations, which ultimately causes the model to overturn an originally correct conclusion.

\subsection{Successful Cases}
\begin{figure}[t]
\centering
\includegraphics[width=1\textwidth]{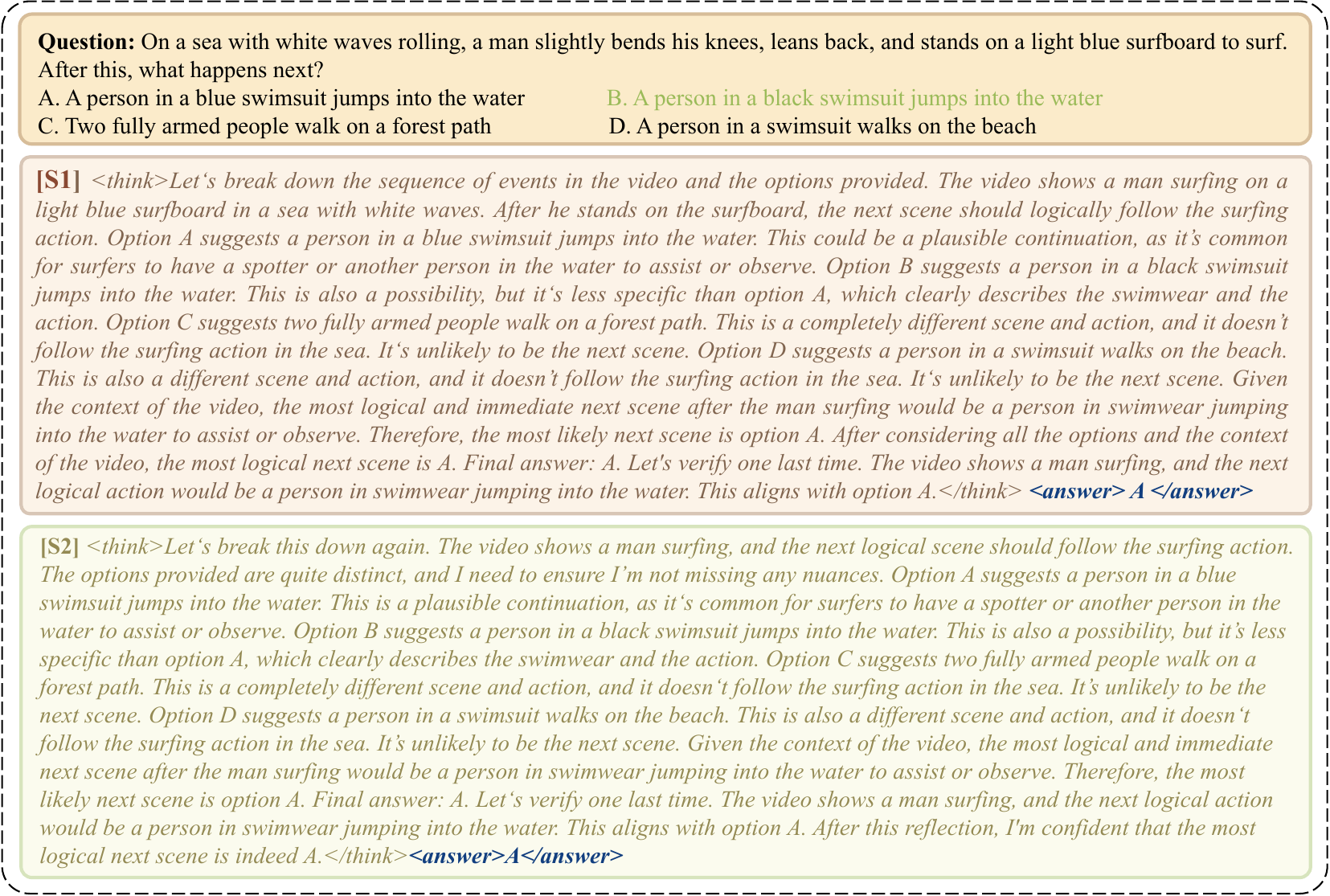}
\caption{
\textbf{Repetition of Initial Errors.} During internal reflection, the model fails to generate new insights and simply repeats the erroneous reasoning of its initial intuition, trapping itself in a hallucination loop.
}
\label{fig:internal_fail_1}
\end{figure}

\begin{figure}[t]
    \centering
    \includegraphics[width=1\textwidth]{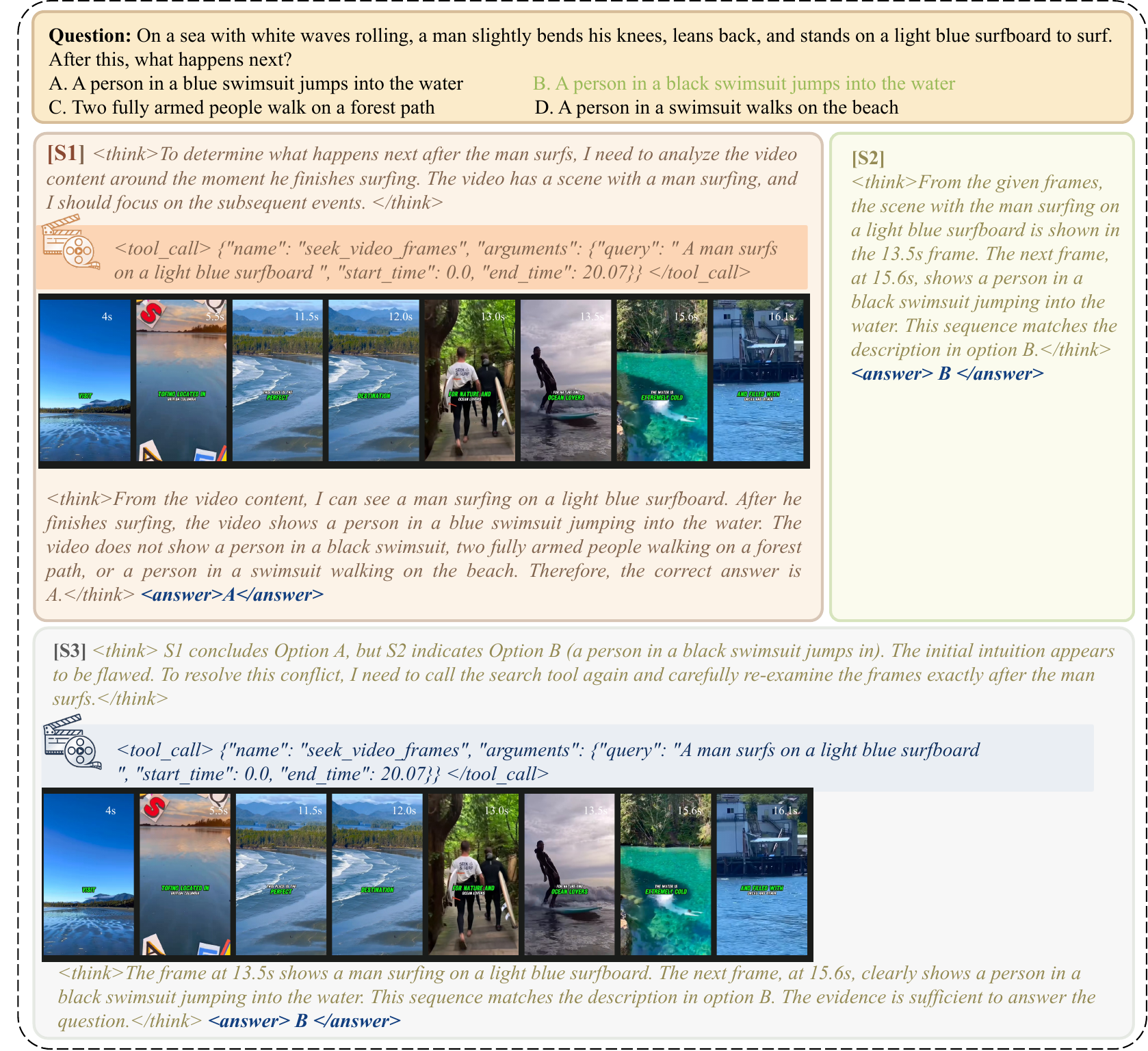}
    \caption{
        \textbf{Evidence-Grounded Hallucination Rectification.} 
        The model resolves the initial visual hallucination by retrieving and grounding its final arbitration strictly on external objective frames.
    }
    \label{fig:case_011}
\end{figure}

\begin{figure}[t]
    \centering
    \includegraphics[width=1\textwidth]{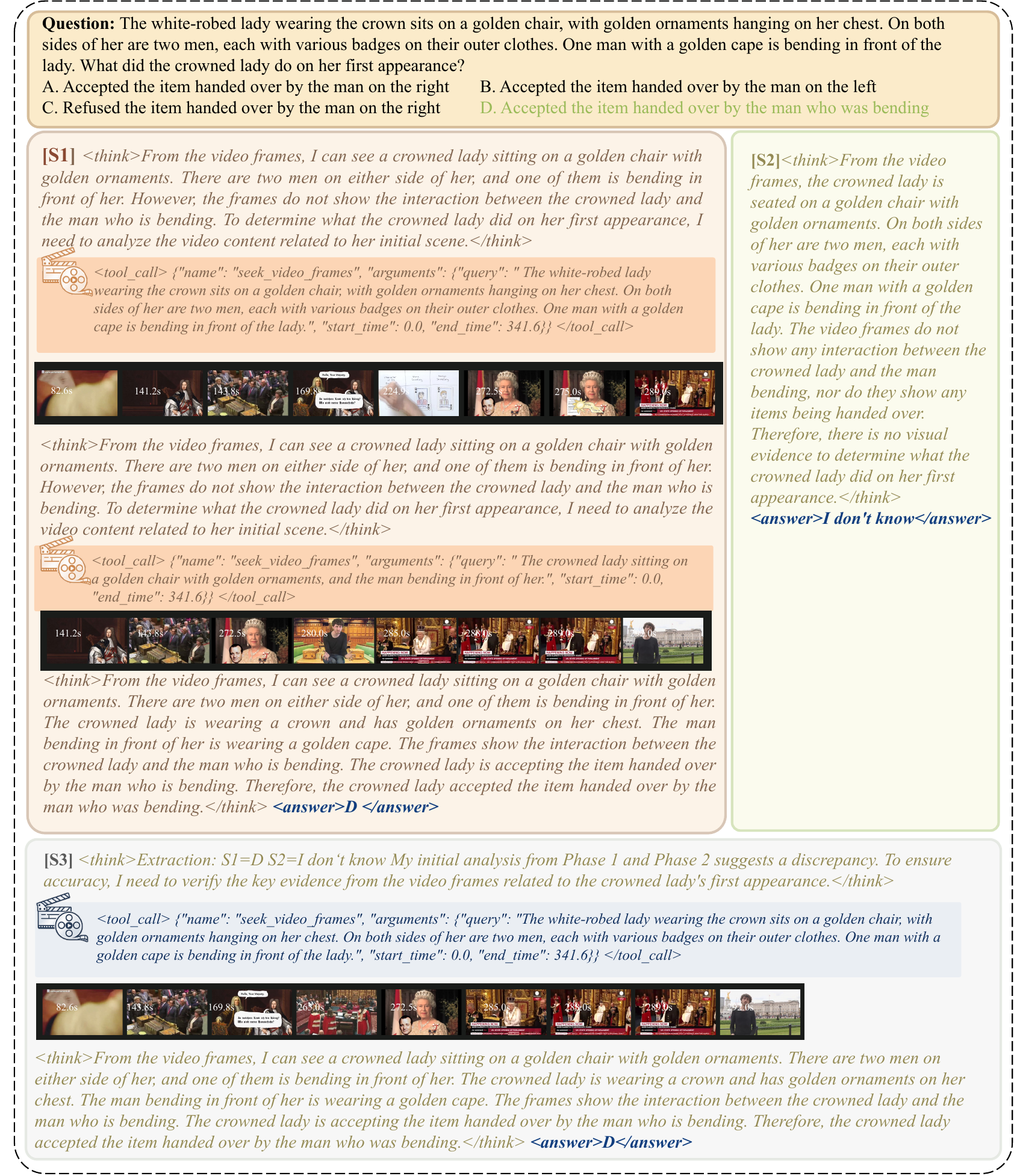}
    \caption{
        \textbf{Abstention and Re-retrieval under Missing Evidence.} 
        When initial retrieval provides insufficient evidence, the model explicitly abstains from guessing and executes a secondary search to secure robust verification.
    }
    \label{fig:case_101}
\end{figure}

\begin{figure}[t]
    \centering
    \includegraphics[width=1\textwidth]{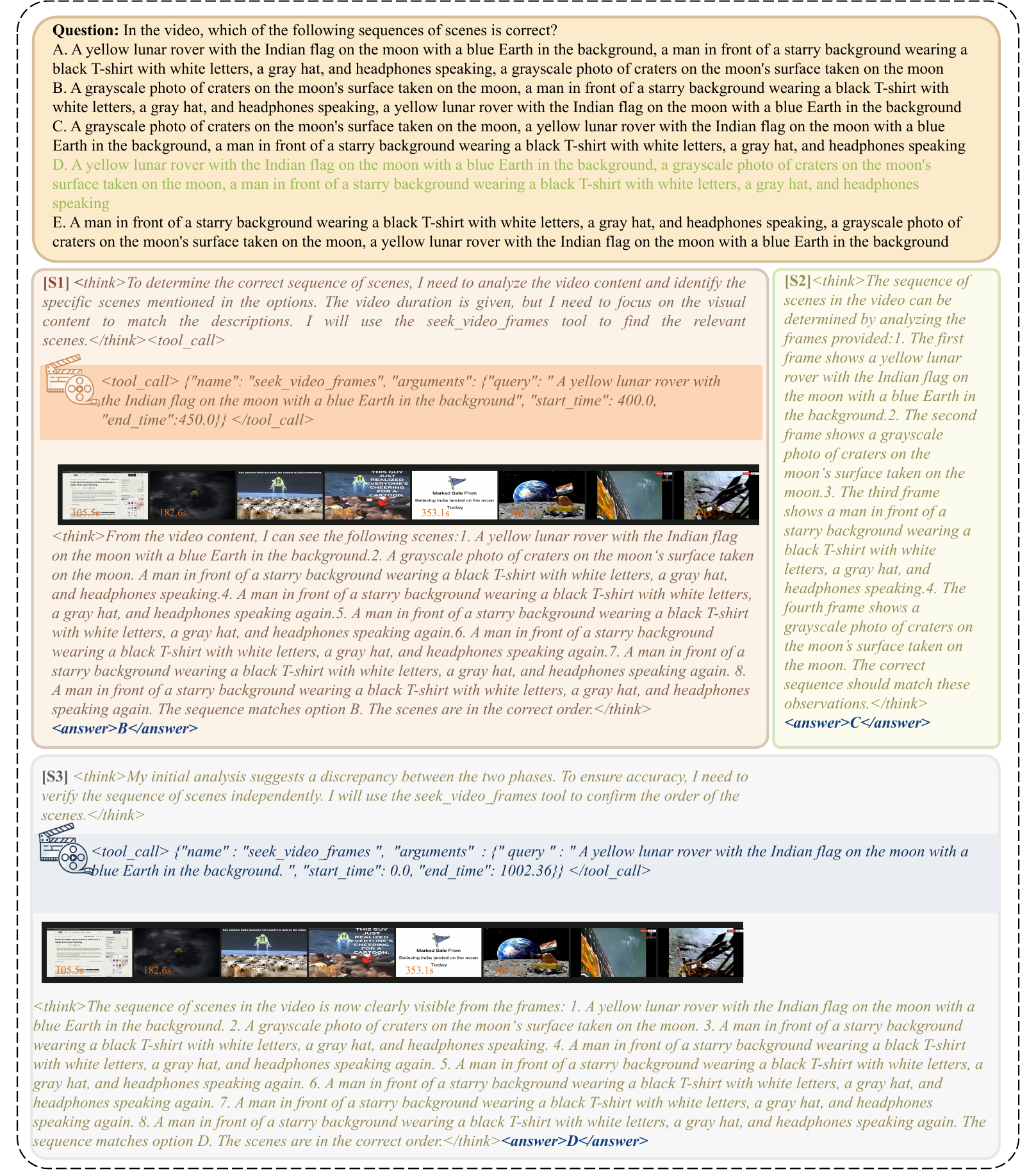}
    \caption{
        \textbf{Global Reconstruction after Local Reasoning Failure.} 
        When prior multi-stage reasoning yields conflicting and incorrect sequences, the arbitrator discards local premises and initiates a global video search to reconstruct the correct timeline.
    }
    \label{fig:case_001}
\end{figure}

Overall, these successful cases demonstrate three representative rectification trajectories. In the first pattern, the model corrects initial visual hallucinations using additionally retrieved external evidence, grounding the final arbitration strictly on objective observations (see Fig.~\ref{fig:case_011}). Notably, this example corresponds to the same question shown in Fig.~\ref{fig:internal_fail_1}, where internal closed-loop reflection fails to revise the incorrect prediction. In contrast, Reflect-R1 successfully resolves the error by retrieving external visual evidence and grounding the final decision in objective observations. In the second pattern, rather than guessing blindly when initial evidence is insufficient, the model explicitly abstains and initiates a secondary retrieval to secure a more robust conclusion (see Fig.~\ref{fig:case_101}). Finally, when local reasoning chains persistently conflict or fail, the model discards unreliable intermediate premises and executes a broader global search to accurately reconstruct the timeline (see Fig.~\ref{fig:case_001}).

\subsection{Failure Cases}
\begin{figure}[t]
\centering
\includegraphics[width=1\textwidth]{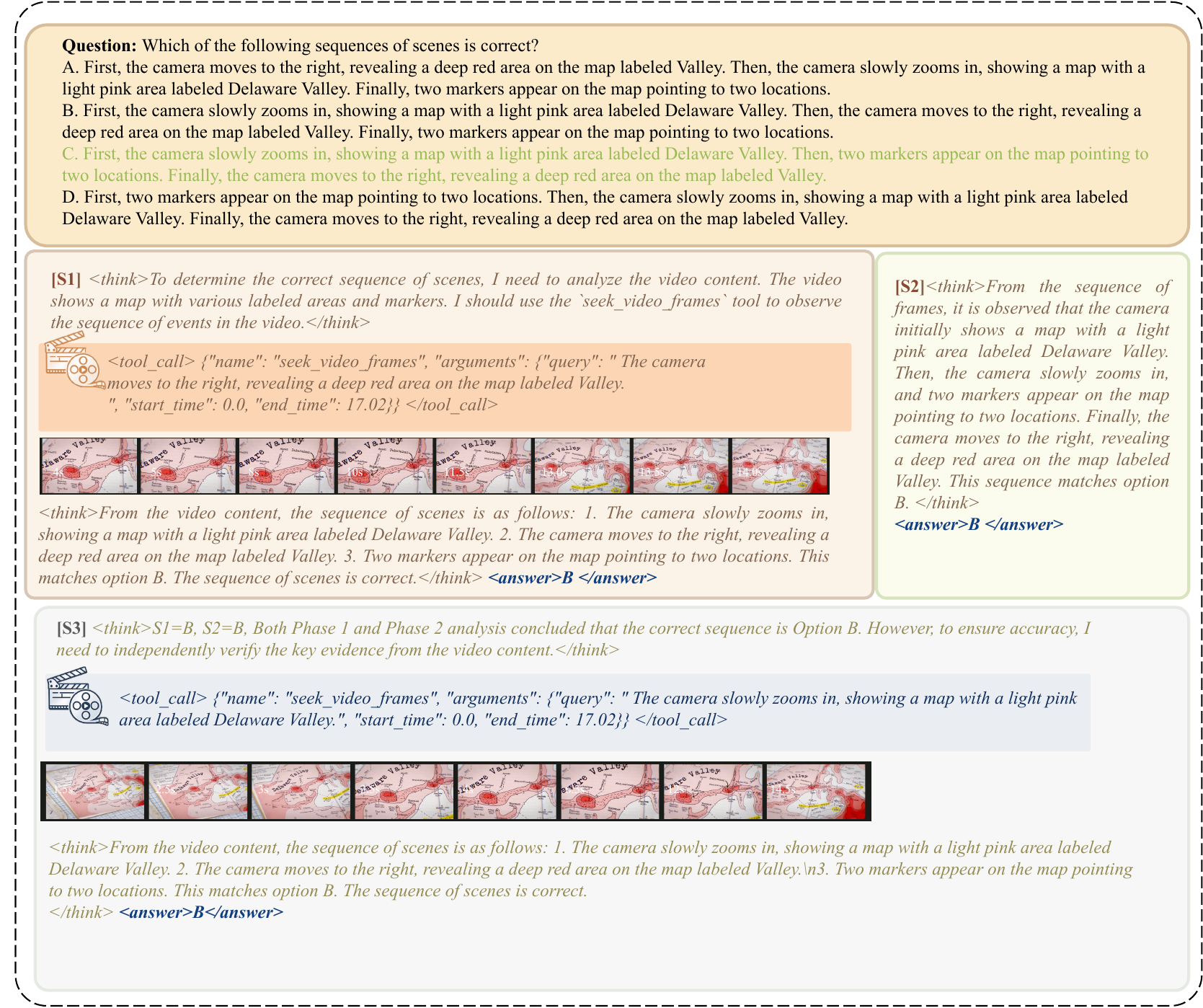}
\caption{
\textbf{Complete Failure.} Both the intuition and verification stages arrive at the same incorrect conclusion. Lacking a conflict signal, the arbitrator adopts the shared error instead of initiating further exploration.
}
\label{fig:case_000}
\end{figure}

\begin{figure}[t]
\centering
\includegraphics[width=1\textwidth]{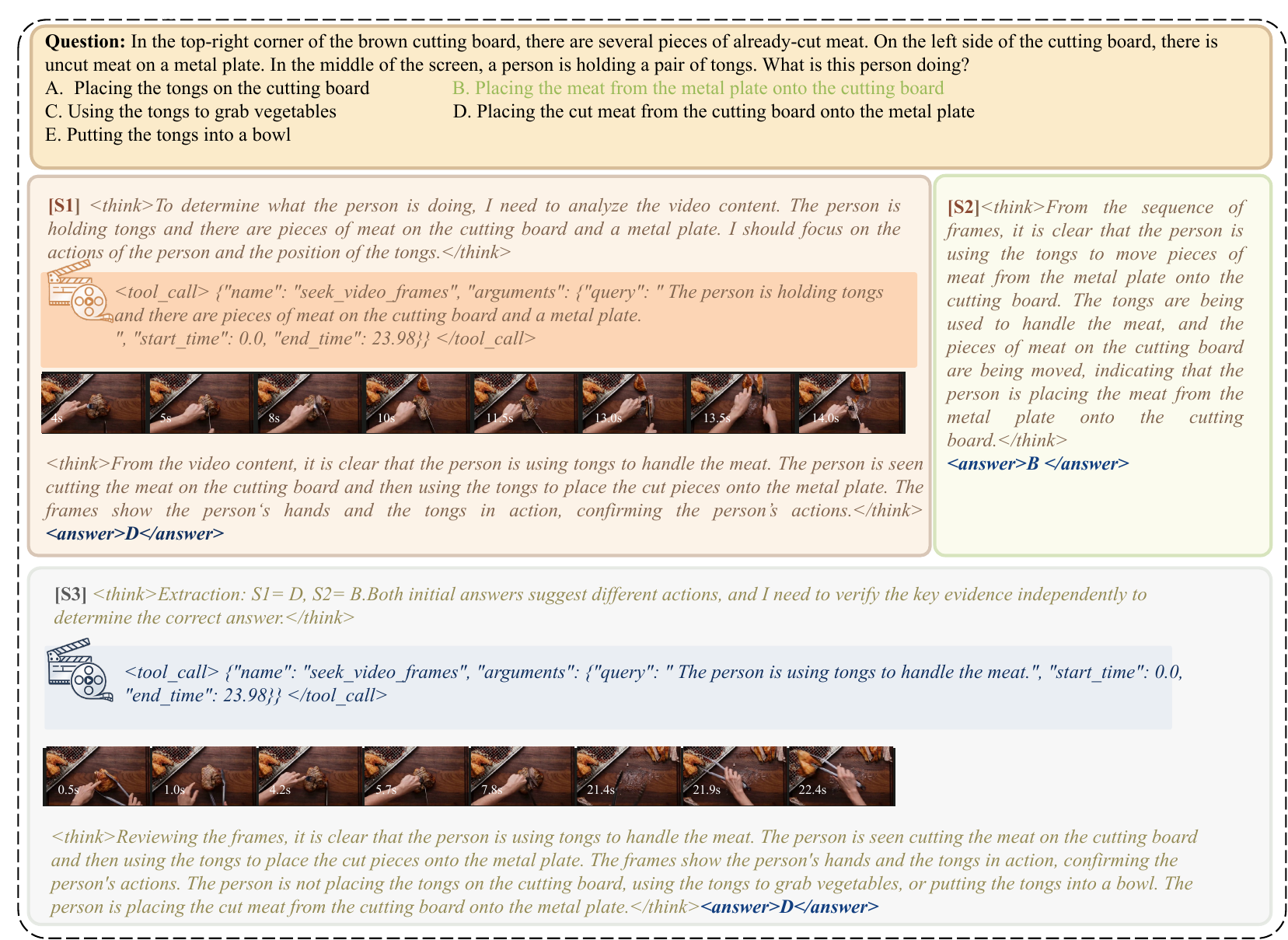}
\caption{
\textbf{Arbitration Failure.} The verification stage successfully identifies the correct answer, but the arbitration stage erroneously reverts to the flawed initial intuition.
}
\label{fig:case_010}
\end{figure}

Although the model demonstrates robust self-correction capabilities, we observe two typical failure patterns that expose the limitations of the current architecture. The first pattern is complete failure (see Fig.~\ref{fig:case_000}). In such highly complex scenarios, both the initial intuition and the independent verification stage arrive at the same incorrect conclusion. Lacking a semantic conflict signal, the arbitration stage fails to initiate a deep rethinking process and directly adopts the erroneous consensus. The second pattern is arbitration failure (see Fig.~\ref{fig:case_010}). In these instances, the verification stage successfully yields the correct answer based on accurately retrieved frames. However, during final arbitration, the model relies excessively on the flawed initial intuition, leaving valid external evidence underutilized. These failure cases indicate that future research must focus on optimizing conflict resolution mechanisms to enhance the reasoning consistency of large models in complex multimodal tasks.






%
%
\ifdefined\ReflectRSkipSupplementReferences
\else
\FloatBarrier
\bibliographystyle{splncs04}
\bibliography{main}
\end{document}
\fi

\end{document}